\documentclass[10pt,twocolumn,letterpaper]{article}

\usepackage{cvpr}
\usepackage{times}
\usepackage{epsfig}
\usepackage{graphicx}
\usepackage{amsmath}
\usepackage{amssymb}
\usepackage{subfigure}
\usepackage{floatrow}
\newcommand{\luwe}{L$^2$UWE}
\usepackage[pagebackref=true,breaklinks=true,letterpaper=true,colorlinks,bookmarks=false]{hyperref}

\cvprfinalcopy 

\def\cvprPaperID{46} 

\ifcvprfinal\fi
\begin{document}

\title{L$^2$UWE: A Framework for the Efficient Enhancement of Low-Light Underwater Images Using Local Contrast and Multi-Scale Fusion}

\author{Tunai Porto Marques, Alexandra Branzan Albu\\
University of Victoria\\
Victoria, British Columbia, Canada\\
{\tt\small tunaip@uvic.ca, aalbu@uvic.ca}
\and
}

\maketitle

\begin{abstract}
   Images captured underwater often suffer from sub-optimal illumination settings that can hide important visual features, reducing their quality. We present a novel single-image low-light underwater image enhancer, \luwe, that builds on our observation that an efficient model of atmospheric lighting can be derived from local contrast information. We create two distinct models and generate two enhanced images from them: one that highlights finer details, the other focused on darkness removal. A multi-scale fusion process is employed to combine these images while emphasizing regions of higher luminance, saliency and local contrast. We demonstrate the performance of \luwe~by using seven metrics to test it against seven state-of-the-art enhancement methods specific to underwater and low-light scenes. Code available at \url{https://github.com/tunai/l2uwe}.
\end{abstract}

\section{Introduction}

The study of underwater sites is important for the environmental monitoring field, as it provides valuable insight regarding their rich ecosystems and geological structures. Sensors placed underwater measure a host of properties, such as physical (e.g., temperature, pressure, conductivity) and biological (e.g., levels of chlorophyll and oxygen); visual data are captured by cameras, and acoustic data are collected with hydrophones. Cabled ocean observatories have installed and maintained various sensor layouts that allow for the continuous monitoring of underwater regions over extended time series. \par

Cabled ocean observatories have captured thousands of hours of underwater videos from fixed and mobile platforms \cite{mdpipaper}. The manual interpretation of these data requires a prohibitive amount of time, highlighting the necessity for semi- and fully-automated methods for the annotation of marine imagery. Mallet \etal~\cite{mallet2014underwater} show that the use of underwater video cameras and their associated software in marine ecology has considerably grown in the last six decades. The efficient interpretation of underwater images requires that they maintain a certain level of \textit{quality} (i.e., contrast, sharpness and color fidelity), which is frequently not possible.\par

Underwater seafloor cameras often can not count on sunlight, prompting the use of artificial means of illumination. These artificial sources are not able to uniformly illuminate the entirety of a scene, and are typically employed in groups. These different and non-uniform lighting setups call for the use of specialized models of atmospheric lighting in image enhancement efforts. Low levels of lighting reduce the quality of visual data because contrast, color and sharpness are deteriorated, making it difficult to detect important features such as edges and textures. Additional challenges posed to the quality of underwater images are related to physical properties of the water medium: \textit{absorption}, which selectively reduces the energy of the propagated light based on its wavelength, and \textit{scattering}. The combined effect of these degradation factors results in images with dark regions, low contrast and hazy appearance. \par

The proposed single-image system, \luwe, offers a novel methodology for the enhancement of low-light underwater images. Its novelty is based on the observation that low-light scenes present particular local illumination profiles that can be efficiently inferred by considering local levels of contrast. We propose two contrast-guided atmospheric illumination models that can harvest the advantages of underwater darkness removal systems such as \cite{mdpipaper}, while preserving colors and enhancing important visual features (e.g., edges, textures). By combining these outputs via a multi-scale fusion process \cite{ancuti2013single} we highlight regions of high contrast, saliency and color saturation on the final result. \par

The performance of \luwe~is compared to that of five underwater-specific image enhancers \cite{berman2017diving,cho2017visibility,drews2013transmission,fu2014retinex,mdpipaper}, as well as two low-light-specific enhancers \cite{guo2016lime, zhang2017fast} using the OceanDark dataset \cite{mdpipaper}. Seven distinct metrics (i.e., UIQM~\cite{panetta2015human}, PCQI~\cite{wang2015patch}, FADE~\cite{choi2015referenceless}, GCF~\cite{matkovic2005global}, \textit{r} and \textit{e}-score~\cite{hautiere2011blind}, SURF features~\cite{bay2008speeded}) show that \luwe~outperforms the compared methods, achieving a significant enhancement in overall image visibility (by reducing low-light regions) and emphasizing the image features (e.g., edges, corners, textures). \par
 \par

\section{Related work}
Sub-section \ref{ss:background} discusses works in three areas of relevance to the development of \luwe: underwater image enhancement, aerial image dehazing and low-light image enhancement. The sub-section that follows, \ref{ss:previous_methods}, summarizes the single image dehazing framework of He \etal~\cite{he2011single}, the multi-scale fusion approach of Ancuti and Ancuti~\cite{ancuti2013single}, and the contrast-guided low-light underwater enhancement system of Marques \etal~\cite{mdpipaper}.   

\subsection{Background}
\label{ss:background}

\textbf{Underwater image enhancement}. Some early approaches that attempted the enhancement of underwater images include: the color correction scheme of Chambah \etal~\cite{chambah2003underwater} that aimed for a better detection of fish, the work of Iqbal \etal~\cite{iqbal2007underwater}, which focused in adjusting the contrast, saturation and intensity to boost images' quality, and the method of Hitam \etal~\cite{hitam2013mixture}, where the equalization of the histogram from underwater images is proposed as a means to achieve enhancement. More recently, Ancuti \etal~\cite{ancuti2012enhancing} introduced a popular framework that derived two inputs (a color-corrected and a contrast-enhanced version of the degraded image), as well as four weight maps that guaranteed that a fusion of such inputs would have good sharpness, contrast and  color distribution.   \par

Multiple underwater image processing methods~\cite{chiang2011underwater,yang2011low,ancuti2016multi,ancuti2017color,peng2015single,drews2013transmission,berman2017diving} make use of aerial dehazing techniques, given that the issues that plague hazy images (absorption and scattering) create outputs that are similar to those captured underwater. Fu \etal~\cite{fu2014retinex} proposed a framework based on the Retinex model that enhances underwater images by calculating their detail and brightness, as well as performing color correction. Berman \etal~\cite{berman2017diving} used the color attenuation and different models of water types to create a single underwater image enhancer. Cho and Kim~\cite{cho2017visibility}, inspired by simultaneous location and mapping (SLAM) challenges, derived an underwater-specific degradation model. Drews \etal~\cite{drews2013transmission} considered only two color channels when using the dehazing approach of He \etal~\cite{he2011single}, adapting this method to underwater scenes. Marques \etal~\cite{mdpipaper} introduced a contrast-based approach inspired in the Dark Channel Prior \cite{he2011single} that significantly improved the quality of low-light underwater images. \par

\textbf{Aerial images dehazing}. Dehazing methods aim for the recovery of the original radiance intensity of a scene and the correction of color shifts caused by scattering and absorption of light by fluctuating particles and water droplets. Initial approaches proposed to address this challenge \cite{narasimhan2000chromatic,schechner2001instant,narasimhan2003contrast, kopf2008deep,treibitz2009polarization} were limited by the need of multiple images (captured under different weather conditions) as input. 
He \etal~\cite{he2011single} proposed a popular method for single-image dehazing that introduced the \textit{Dark Channel} and the \textit{Dark Channel Prior} (DCP), which allow for the estimation of the transmission map and atmospheric lighting of a scene, both important parameters of the dehazing process (as detailed in sub-section \ref{ss:previous_methods}). Comparative results~\cite{ancuti2016d,ren2016single,alharbi2016research, mdpipaper} attest to the performance of this method in scenarios where only a single hazy image is~available. Numerous underwater-specific enhancement frameworks  \cite{mdpipaper,drews2013transmission,carlevaris2010initial, chiang2011underwater, galdran2015automatic, lu2015contrast} are based on variations of the DCP. Recently a number of data-driven methods \cite{cai2016dehazenet, ren2016single, zhang2017learning, li2017aod, ren2018gated, zhang2018density} focused on the use of Convolution Neural Networks (CNNs) to train systems capable of efficiently performing the dehazing/image recovering task. However, these systems typically require time-consuming processes of data gathering and curation, hyper-parameter tuning, training, and evaluation.\par
\textbf{Low-light image enhancement}. Dong \etal~\cite{dong2011fast} observed that dark regions in low-light images are visually similar to haze in their inverted versions. The authors proposed to use the DCP-based dehazing method to remove such haze. Ancuti \etal \cite{ancuti2016night} proposed to use a non-uniform lighting distribution model and the DCP-based dehazing method to generate two inputs (an additional input is the Laplacian of the original image), followed by a multi-scale fusion process. Zhang \etal~\cite{zhang2017fast} introduced the maximum reflectance prior, which states that the maximum local intensity in low-light images depends solely on the scene illumination source. This prior is used to estimate the atmospheric lighting model and transmission map of a dehazing process (refer to sub-section \ref{ss:previous_methods} for details). Guo \etal~\cite{guo2016lime} proposed LIME, where the atmospheric lighting model is also initially constructed by finding the maximum intensity throughout the color channels. The LIME framework refines this initial model by introducing a structure prior that guarantees structural fidelity to the output, as well as smooth textural details. Data-driven approaches were also proposed for the enhancement of low-light images \cite{Jiang_2018,wei2018deep,shen2017msr,jiang2019enlightengan}. These methods employ CNNs to extract features from datasets composed of  low-light/normal-light pairs of images (\cite{jiang2019enlightengan} requires only the degraded images), and then train single-image low-light enhancement frameworks. \par

\subsection{Previous Works Supporting the Proposed Approach}
\label{ss:previous_methods}

\textbf{Dark Channel Prior-based dehazing of single images}. Equation \ref{eq:hazyimageformation} describes the formation of a hazy image $I$ as the sum of two additive components, \textit{direct attenuation} and \textit{airlight}. 

\begin{equation} \label{eq:hazyimageformation}
I(x) = J(x)t(x) + A_{\infty}(1-t(x))
\end{equation}

where $J$ represents the haze-less version of the image, $x$ is an spatial location, transmission map $t$ indicates the amount of light that reaches the optical system and $A_{\infty}$ is an estimation of the global atmospheric lighting. The first term, direct attenuation $D(x) = J(x)t(x)$, indicates the attenuation suffered by the scene radiance due to the properties of the medium. The second term, airlight $V(x) = A_{\infty}(1-t(x))$, is due to previously scattered light and may result in color shifts on the hazy image. The~goal of dehazing efforts is to find the haze-free version of the image ($J$) by determining $A_{\infty}$ and $t$. \par

He \etal~\cite{he2011single} introduced the \textit{Dark Channel} and the \textit{Dark Channel Prior}, which can be used to infer atmospheric lighting $A_{\infty}$ and to derive a simplified formula for the calculation of $t$. The dark channel is described in Equation \ref{eq:dc}. 

\begin{equation} \label{eq:dc}
J^{dark}(x) = \min_{y \in \Omega(x)}(\min_{c \in \{r,g,b\}} I^c(y))
\end{equation}

where $x$ and $y$ represent two pairs of spatial coordinates in the dark channel $J^{dark}$ and in the hazy image $I$ (or any other arbitrary image), respectively. The intensity of each pixel in the single-channel image $J^{dark}$ is the lowest value between the pixels inside patch $\Omega$ in $I^{c}$, where $c \in \{R, G, B\}$. The DCP is an empirical observation stating that pixels from non-sky regions have at least one significantly low intensity across the three color channels. Thus the dark channel is expected to be mostly dark (i.e., intensities closer to 0).\par
A single three-dimensional global atmospheric lighting vector $A^{c}_{\infty}$ ($c \in \{R, G, B\}$) can be inferred by looking at the 0.1\% \cite{he2011single} or 0.2\% \cite{mdpipaper} brightest pixels in the dark channel (which represent the most haze-opaque regions from input), then considering the brightest intensity pixels in these same spatial coordinates from the \textit{input} image $I$. Ancuti \etal \cite{ancuti2016night} observed that a single global value might not properly represent the illumination of low-light scenes, thus proposing the estimation of local atmospheric light intensities $A^{c}_{L\infty}$ inside patches $\Psi$ following Equation \ref{eq:local_atml}. 

\begin{equation} \label{eq:local_atml}
A^{c}_{L\infty}(x) = \max_{y \in \Psi(x)}(\min_{z \in \Omega(y)}(I^{c}(z)))
\end{equation}

where $x$, $y$ and $z$ represent, respectively, spatial coordinates in the local atmospheric image, ``minimum'' image $I_{\min}(z) = \min_{z \in \Omega(y)}(I(z))$, and hazy image $I$. For each color channel $c \in \{R, G, B\}$, the local atmospheric lighting $A^{c}_{L\infty}$ is calculated by first finding $I^{c}_{\min}$, which represents the minimum intensities inside patch $\Omega$ on $I^{c}$, and then calculating the maximum intensities inside a patch $\Psi$ on $I^{c}_{\min}$. By arguing that the influence of lighting sources goes beyond patch $\Omega$, Ancuti \etal \cite{ancuti2016night} used patch $\Psi$ with twice the size of  $\Omega$. The DCP is used in \cite{he2011single} to derive Equation \ref{eq:transmissionmap} for the calculation of the transmission map $t$. 

\begin{equation} \label{eq:transmissionmap}
t(x) = 1 - \omega \min_{y \in \Omega(x)} (\min_{c \in \{r,g,b\}} \frac{I^c(y)}{A^c_{\infty}} )
\end{equation}

where $A^c_{\infty}$ indicates the atmospheric lighting in the range $[0,255]$, resulting in a normalized image ($\frac{I^c(y)}{A^c_{\infty}}$) ranging from $[0,1]$. The constant $\omega$ $(0\leq \omega \leq1)$ preserves a portion of the haze, generating more realistic output. Note that for local estimation of atmospheric lighting, $A^c_{\infty}$ would be substituted by $A^{c}_{L\infty}$ in Equation \ref{eq:transmissionmap}. The haze-less version of the image, $J(x)$, is obtained using Equation \ref{eq:radiancerecovery} \cite{he2011single}. 

\begin{equation} \label{eq:radiancerecovery}
J^{c}(x) = \frac{I^{c}(x)-A^c_{\infty}}{\max{(t(x),t_0)}}+A^c_{\infty}
\end{equation}

Since the direct attenuation $D(x) = J(x)t(x)$ can be close to zero when $t(x)\approx0$, \cite{he2011single} adds the $t_0$ term as a lower bound to $t(x)$, effectively preserving small amounts of haze in haze-dense regions of $I$.   

\textbf{Contrast-guided approach for the enhancement of low-light images}. Marques \etal \cite{mdpipaper} observed and addressed three problems that arrive from the use of single-sized patches $\Omega$ throughout the image dehazing process: 1) small patch sizes would result in oversaturation of the radiance from the  recovered scene (non-natural colors); 2) large patch sizes better estimate and eliminate haze, but since they consider that the transmission profile (i.e., amount of light that reaches a camera) is constant inside patch $\Omega$, undesired halos might appear around intensity discontinuities and 3) a single patch size is typically not optimal for images of varying scales.\par

Marques \etal \cite{mdpipaper} argues that homogeneous regions of an image possess lower contrast, and that the assumption that their transmission profile is approximately the same holds for patches $\Omega$ of varying sizes (in particular, from $3\times3$ up to $15\times15$). In these scenarios, the use of larger patch sizes generates darker dark channels (strengthening the DCP) and are, therefore, preferred. For regions with complex content (i.e., intensity changes), \cite{mdpipaper} uses smaller patch sizes to capture the local transmission profile nuances. To determine the correct patch size for each pixel in image $I$, \cite{mdpipaper} introduced the \textit{contrast code image} ($CCI$), whose calculation is summarized in Equation \ref{eq:CCI}.   
\begin{equation} \label{eq:CCI}
CCI(x) = \arg \min_{i} [\sigma(\Omega_{i}(x))]
\end{equation}
where $\Omega_{i}(x) \in I$ represents a square patch of size $(2i~+~1)\times(2i~+~1)$ centered at the pair of spatial coordinates $x$, $i=\{1,2,...,7\}$, and $\sigma$ represents the standard deviation between intensities inside of patch $\Omega_{i}$ (considering the three color channels). Therefore, the $CCI$ is populated by $i$ (referred to as code $c$ in \cite{mdpipaper}), which indicates the size of the patch $\Omega$ that generated the smallest $\sigma$ in each pixel location $x$. A variable named $tolerance$ is also introduced to incentivize the use of larger patch sizes by changing the measured values of $\sigma$ for different $i$.\par

The $CCI$ is then used to calculate the transmission map and dark channel. For a pixel location $x$, one would use patches of size $(2c~+~1)\times(2c~+~1)$ (where $c = CCI(x)$) instead of fixed-size patches. This constrast-guided approach significantly mitigates the three aforementioned problems. 

\textbf{Multi-scale fusion for image enhancement}. The work of Ancuti \etal \cite{ancuti2013single} proposes to perform dehazing by first calculating two inputs $\mathcal{I}^{k}$ ($k=\{1,2\}$) from the original image: a white-balanced and a contrast-enhanced version of it. Then the authors introduced the now-popular multiscale fusion process, where three \textit{weight maps} containing important features of each input $\mathcal{I}^{k}$ are calculated: 1) \textbf{Luminance $\mathcal{W}_L^k$}: responsible for assigning high values to pixels with good visibility, this weight map is computed by observing the deviation between the R, G and B color channels and luminance $L$ (average of pixel intensities at a given location) of the input; 2) \textbf{Chromaticity $\mathcal{W}_C^k$}: controls the saturation gain on the output image, and can be calculated by measuring the standard deviation across each color channel from the input; 3) \textbf{Saliency $\mathcal{W}_S^k$}: in order to highlight regions with greater saliency, this weight map is obtained by subtracting a Gaussian-smoothed version of the input by its mean value (method initially proposed by Achanta \etal \cite{achanta2009frequency}). 
Aiming to minimize visual artifacts introduced by the simple combination of the weight maps, \cite{ancuti2013single} uses a multiscale fusion process where a Gaussian pyramid is calculated for the normalized weight map $\bar{\mathcal{W}}^k$ (i.e., per-pixel division between the multiplication of all three weights and their sum) of each input, while the inputs $\mathcal{I}^{k}$ themselves are decomposed into a Laplacian pyramid. Given that the number of levels from both pyramids is the same, they can be independently fused using Equation \ref{eq:multi_scale_fusion}.
\begin{equation} \label{eq:multi_scale_fusion}
\mathcal{R}_{l}(x) = \sum_{k} G_{l} \{\bar{\mathcal{W}}^{k}(x)\} L_{l} \{\mathcal{I}_{k}(x)\}
\end{equation}
where $l$ indicates the number of levels (typically 5), $L\{\mathcal{I}\}$ represents the Lapacian of $\mathcal{I}$, and $G\{\bar{\mathcal{W}}\}$ denotes the Gaussian-smoothed version of $\bar{\mathcal{W}}$. The fused result is a sum of the contributions from different layers, after the application of an appropriate upsampling operator. \par
The authors applied the multi-scale fusion with different strategies in other works, for example by changing the inputs to gamma-corrected and sharpened versions of a white-balanced underwater image \cite{ancuti2017color} or by calculating new weight maps (e.g., local contrast weight map \cite{ancuti2016night}). 

\section{Proposed approach}

Although providing good enhancement results for low-light underwater images, the contrast-guided approach of \cite{mdpipaper} oversimplifies the scene atmospheric lighting  using a single 3-channel value $A_{\infty}$. This common practice, which assumes a mostly white-colored lighting source, works well with aerial hazy images under sunlight \cite{he2011single}, but fails to represent low-light scenarios properly \cite{ancuti2016night}, since those can present non-homogeneous and non-white illumination. This misrepresentation in \cite{mdpipaper} results in images with regions that suffer an excess of darkness removal, generating outputs that are overly bright and have a washed-out appearance, often belittling intensity discontinuities (e.g., edges, textures) that could represent important visual features in other computer vision-based applications. This undesirable phenomenon results in Marques \etal method's low $e$-score \cite{hautiere2011blind} and count of SURF~\cite{bay2008speeded} features (results presented in \cite{mdpipaper}). \par

\begin{figure*}[!htbp]
\centering
\includegraphics[width=0.9\linewidth]{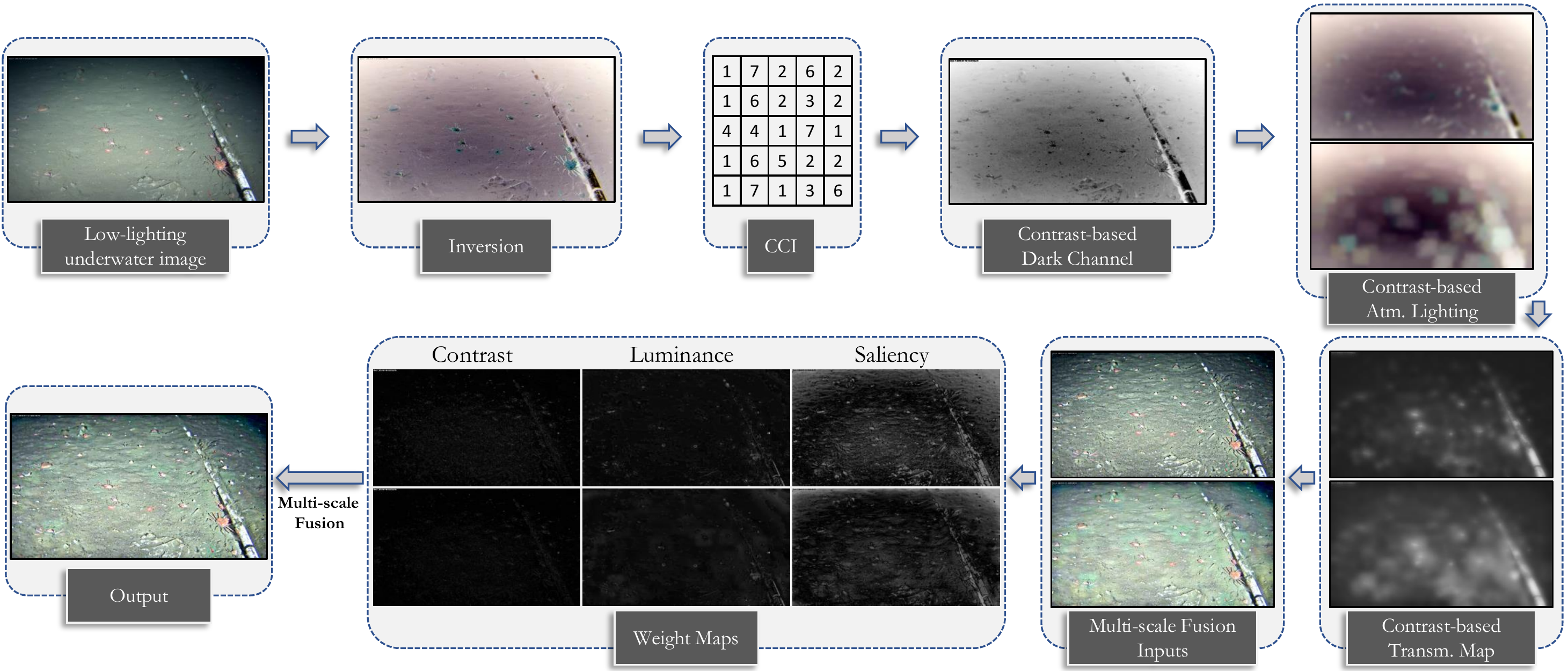} 
\vspace{-0.5em}
\caption{Pipeline of \luwe. First, the $CCI$ \cite{mdpipaper} is calculated in the inverted version of the image. The contrast-based Dark Channel is then used to derive two atmospheric lighting models that consider local illumination. Two transmission maps generate the inputs of a multi-scale fusion process. Three weight maps are calculated for each input, which are then combined to offer the framework's output.}

\label{fig:overall}
\end{figure*}

With \luwe~we propose a better image enhancement mechanism by deriving more realistic models for underwater illumination. We consider local contrast information as a guiding parameter to derive lighting distribution models under two distinct ``lenses'': one very focused (i.e., using smaller local regions) that captures the finer details of the original image, and a second, wider one, which considers larger local regions to create brighter models. Each model drives a different dehazing process, and the outputs are combined using a multi-scale fusion approach. This fusion strategy preserves both the details and darkness removal obtained with the two lighting models. As a result, the output of \luwe~drastically reduces the amount of darkness from the original images while highlighting their intensity changes. Figure \ref{fig:overall} details the computational pipeline of \luwe.\par

\subsection{Contrast-aware local atmospheric lighting models for low-light scenes}
\label{ss:atm_lighting}
Using parts of the dark channel to derive a single \textit{global estimate} $A_{\infty}$ for underwater images creates overly-bright and washed-out results. The underlying assumption in \cite{mdpipaper} that the lighting in inverted input images is mostly white is not reasonable for underwater scenes: see the ``inversion'' step of Figure \ref{fig:overall}, where the lighting presents non-white colors. Given that the images from the OceanDark dataset \cite{mdpipaper} are captured using artificial (and often multiple) lighting sources, a single global estimate $A_{\infty}$ can not properly model their non-uniform nature. \luwe~addresses these problems by calculating the atmospheric lighting for each color channel, and by considering the $CCI$ code at each spatial position $x$ when determining \textit{local estimates} of lighting. This approach is similar to that described in Equation~\ref{eq:local_atml} \cite{ancuti2016night}, but instead of using a fixed-size patch $\Psi$, we introduce the contrast-guided patch $\Upsilon$ for lighting calculation.\par
We observed that regions with high contrast (i.e., lower code $c$ in the $CCI$) should have their local lighting component modeled by considering a larger $\Upsilon$, based on the heterogeneous influence that illumination sources place on them. On the other hand, since regions with lower contrast (i.e., higher codes $c$ in the $CCI$) are illuminated in a roughly homogeneous manner, only smaller patches $\Upsilon$ need to be studied to properly model local illumination. We offer a formalization of this reasoning in Equation \ref{eq:mult_factor}, which specifies the relationship between $CCI$ code $c$ and the size of the lighting square patch $\Upsilon$ used in our local atmospheric lighting model calculation.
\begin{equation} \label{eq:mult_factor}
S_{\Upsilon}(m,c) = 3m - [\frac{m}{3}(c-1)]
\end{equation}
where $m$ represents an arbitrary multiplication factor and $c=\{1,2,...,7\}$ represents code $c$ in the $CCI$ at a certain position. Parameter $m$ has a multiplicative effect on the contrast-guided patch, but an offset is also added to progressively constrain it based on the patch size: smaller patches will be more influenced than larger patches. For example, for $m=15$ and a position $x$ where $CCI(x)=1$, patch $\Upsilon(x,m)$ would be of dimensions $S_{\Upsilon}(15,1)\times S_{\Upsilon}(15,1)$, or $45\times45$. Similarly, for $m=15$ and $CCI(x)=5$ (lower contrast), patch $\Upsilon(x,m)$ would be of dimensions $25\times25$.\par
With the use of contrast-aware patch $\Upsilon$ and multiplication factor $m$, we define the local, contrast-guided atmospheric intensity $A^{c}_{LCG\infty}$ for a color channel $c$ as:  
\begin{equation}
    \label{eq:atm_LCG}
    A^{c}_{LCG\infty}(x,m) = \max_{y \in \Upsilon(x,m)}(\min_{z \in \Omega(y)}(I^{c}(z)))
\end{equation}

\begin{figure}[!htbp]
\centering
   \includegraphics[width=0.32\linewidth]{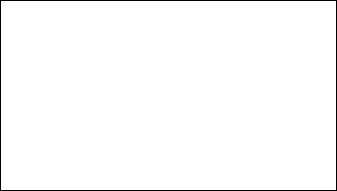}
   \includegraphics[width=0.32\linewidth]{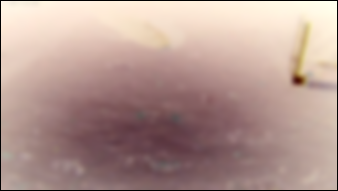}
   \includegraphics[width=0.32\linewidth]{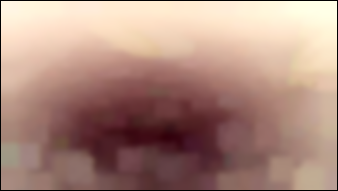}
\vspace{-0.5em}
   \caption{Different approaches for atmospheric lighting estimation on inverted OceanDark sample. \textbf{Left:} a single 3-channel value ($A_\infty$), implying that the illumination is homogeneous and roughly white. \textbf{Center:} contrast-aware local lighting estimation $A^{c}_{LCG\infty}$. \textbf{Right:} Gaussian-filtered version of $A^{c}_{LCG\infty}$, preventing the creation of halos in the output.}
\label{fig:atm_lighting}
\end{figure}

The main distinction between $A^{c}_{LCG\infty}(x,m)$ (Equation~\ref{eq:atm_LCG}) and $A^{c}_{L\infty}(x)$ (Equation~\ref{eq:local_atml}) is that the former uses contrast-aware patches $\Upsilon(x,m)$, while the latter uses fixed-size patches $\Psi(x)$. Thus we refer the reader to Equation~\ref{eq:local_atml} and its discussion on sub-section \ref{ss:previous_methods}.  Since $A^{c}_{LCG\infty}$ is calculated for each color channel $c$, by maximizing contrast-aware patch $\Upsilon(x,m)$ one is actually determining a local position $y$ where the radiance for a certain color channel $c$ is maximum in the $I^c_{min}$ (i.e., a dark channel specific for color $c$). Figure \ref{fig:atm_lighting} compares the different atmospheric lighting models obtained using $A_{LCG\infty}$ and $A_{\infty}$. The $A_{LCG\infty}$ is filtered using a Gaussian kernel with $\sigma=10$ to prevent the creation of abrupt, square-like intensity discontinuities (``halos'') when normalizing the input image by the atmospheric lighting (see Equation~\ref{eq:transmissionmap}). Differently from $A_{\infty}$, $A_{LCG\infty}$ captures the color characteristics of the illumination, as well as its local distribution throughout the image.        

\subsection{Fusion process}

\begin{figure}[!htbp]
\centering
   \includegraphics[width=0.32\linewidth]{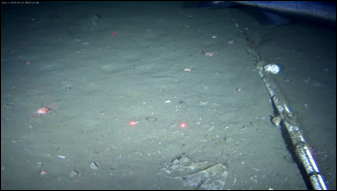}
   \includegraphics[width=0.32\linewidth]{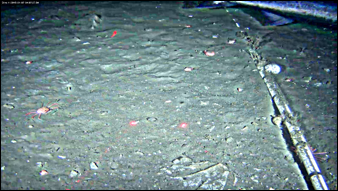}
   \includegraphics[width=0.32\linewidth]{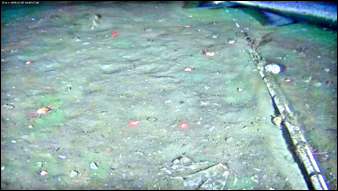}

\vspace{-0.5em}  
   \caption{Multi-scale process inputs. \textbf{Left:} original OceanDark sample. \textbf{Center:} first input, an image enhanced using the contrast-guided $CCI$ and $A_{LCG\infty}$ with $m=5$. \textbf{Right:} second input, an image enhanced using the contrast-guided $CCI$ and $A_{LCG\infty}$ with $m=30$.    }
\label{fig:inputs_10_30}
\end{figure}

The choice of parameter $m$ in Equations~\ref{eq:mult_factor} and \ref{eq:atm_LCG} determines the size of local window $\Upsilon$ and therefore the radius of influence from each illumination source. In other words, higher $m$ creates brighter lighting models. While this generates output with less darkness, it also might apply an excess of radiance correction (i.e., bringing intensities close to saturation) in regions of the image and hide important intensity changes (as previously discussed about \cite{mdpipaper}). In order to harvest the advantages of both choices, we derive two $A_{LCG\infty}$: one with $m=5$ and the other choosing $m=30$. These lighting models are used with Equation~\ref{eq:transmissionmap} to determine two transmission maps. These maps are filtered using a Fast Guided filter \cite{he2013guided}, and finally used to recover two haze-less versions of the original image (Equation~\ref{eq:radiancerecovery}). The image enhancement results obtained using each $A_{LCG\infty}$ serve as inputs $\mathcal{I}^{k}$ ($k=\{1,2\}$) to a multi-scale fusion process. Figure \ref{fig:inputs_10_30} shows two inputs generated with $m=\{5,30\}$ for an OceanDark sample. 

By fusing the inputs generated from different $A_{LCG\infty}$, we preserve two important aspects of the enhanced images: the efficient darkness removal of the input obtained with $m=30$ (Figure~\ref{fig:inputs_10_30} right) and the edges, textures and overall intensity changes of the input generated with $m=5$ (Figure~\ref{fig:inputs_10_30} center). Experiments evaluating different pairs of $m$ values yielded the best performance when using $m=5$ and $m=30$. We also performed tests using a higher number of inputs, but similar results were obtained as for fusing the inputs corresponding to $m=5$ and $m=30$, at the expense of a higher computational complexity.  \par


In order to properly combine the two inputs, we calculate three weight maps from each of them: saliency, luminance and local contrast. These weight maps guarantee that regions in the inputs that have high saliency and contrast, or that possess edges and texture variations will be emphasized in the fused output \cite{ancuti2013single,ancuti2016night}.\par

As discussed in sub-section \ref{ss:previous_methods}, the \textbf{saliency weight map} is calculated by subtracting a Gaussian-smoothed version of input $k$, $\mathcal{I}^{G_s}_{k}$, by the mean intensity value of this same input, $\mathcal{I}_{k}^{\mu}$  (constant for each input), as detailed in Equation \ref{eq:saliency_wm}.

\begin{equation}
    \mathcal{W}_S^k(x)=\parallel \mathcal{I}^{G_s}_{k}(x)-\mathcal{I}_{k}^{\mu}  \parallel
    \label{eq:saliency_wm}
\end{equation}

where $x$ represents a spatial location of input $k$ and $\mathcal{I}^{G_s}_{k}$ is obtained with a $5\times5~(\frac{1}{16}[1,4,6,4,1])$ Gaussian kernel.\par

Considering that saturated colors present higher values in one or two of the $R, G, B$ color channels \cite{ancuti2013single}, we use Equation \ref{eq:luminance_wm} to calculate the \textbf{luminance weight map} $\mathcal{W}_L^k$.

\begin{equation}
    \mathcal{W}_L^k = \sqrt{\frac{1}{3}[(R^k-L^k)^2+(G^k-L^k)^2+(B^k-L^k)^2}]
\label{eq:luminance_wm}
\end{equation}

where $L^{k}$ represents, at each spatial position, the mean of R, G, B intensities for input $k$. $R^{k}, G^{k}$ and $B^{k}$ are the three color channels of input $\mathcal{I}^{k}$.\par

The \textbf{local contrast weight map} $\mathcal{W}_{LCon}^k$, also used in \cite{ancuti2016night}, is responsible for highlighting regions of input $\mathcal{I}^{k}$ where there is more local intensity variation.  We calculate it by applying a $ \frac{1}{8}\big[\begin{smallmatrix}
    -1 & -1 & -1\\
    -1 & 8 & -1\\
    -1 & -1 & -1\\
\end{smallmatrix}\big]$ Laplacian kernel on $L^{k}$. \par

\begin{figure}[!htbp]
\centering
\subfigure[]
    {\includegraphics[width=0.325\linewidth]{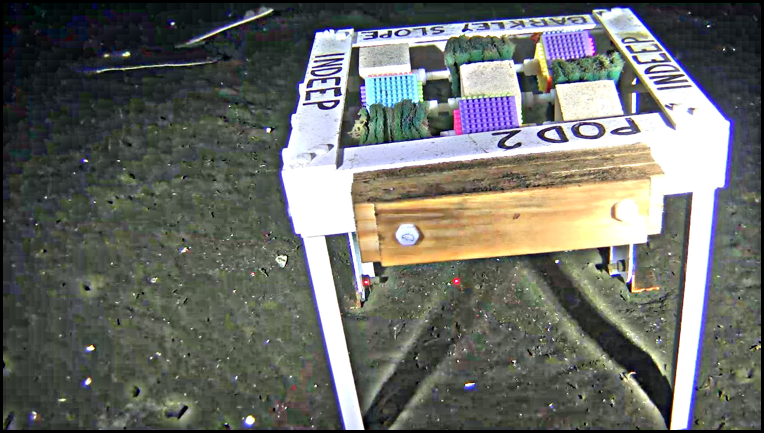}}\hspace{-0.1em}\vspace{-0.3em}
\subfigure[]
    {\includegraphics[width=0.325\linewidth]{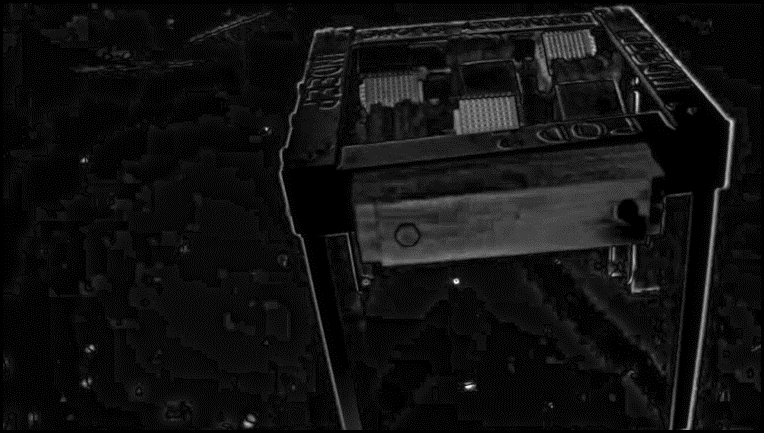}}\hspace{-0.1em}\vspace{-0.3em}
\subfigure[]
    {\includegraphics[width=0.325\linewidth]{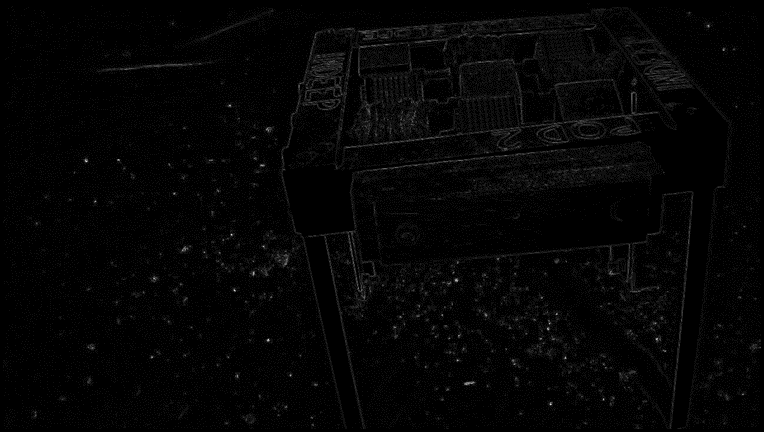}}
    \vspace{-0.3em}
\subfigure[]
    {\includegraphics[width=0.325\linewidth]{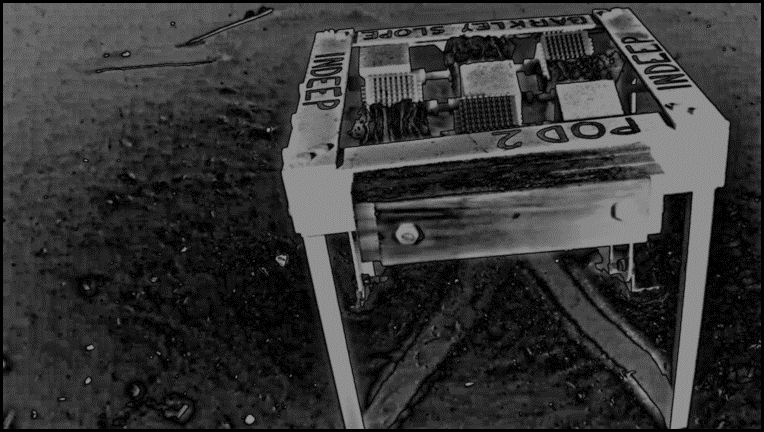}}\hspace{-0.1em}
\subfigure[]
    {\includegraphics[width=0.325\linewidth]{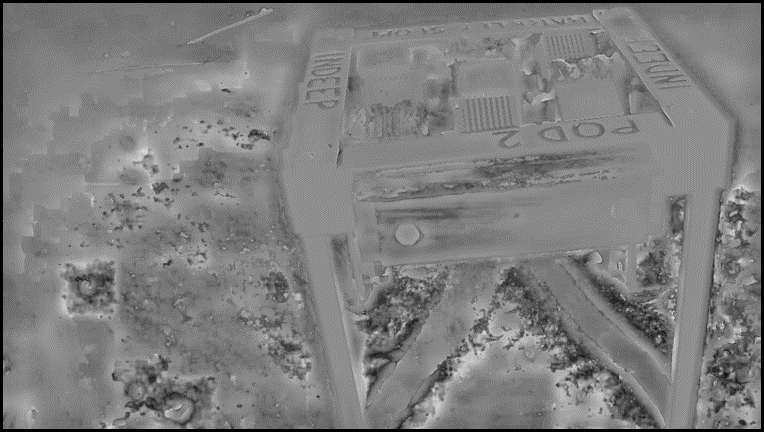}}
   
\vspace{-1em}   

   \caption{Weight maps of an input $\mathcal{I}^{k}$. \textbf{(a)} Input image. \textbf{(b)} Luminance weight map $\mathcal{W}_L^k$. \textbf{(c)} Local contrast weight map $\mathcal{W}_{LCon}^k$. \textbf{(d)} Saliency weight map $\mathcal{W}_S^k$. \textbf{(e)} Normalized weight map $\bar{\mathcal{W}}^k$.       
   }
\label{fig:weight_maps}
\end{figure}

The three weight maps ($\mathcal{W}_{LCon}^k$, $\mathcal{W}_L^k$, $\mathcal{W}_S^k$) are combined into a normalized weight map $\bar{\mathcal{W}}^k$, from which a $5$-level Gaussian pyramid $G\{\bar{\mathcal{W}}^k\}$ is derived. Our choice of using Gaussian pyramids is based on their efficacy in representing weight maps, as demonstrated by Ancuti\etal~\cite{ancuti2013single}. Figure \ref{fig:weight_maps} illustrates the three weight maps calculated for one input image, as well as their equivalent normalized weight map.\par

\begin{figure}[!htbp]
\centering
\subfigure[]
{\includegraphics[width=0.46\linewidth]{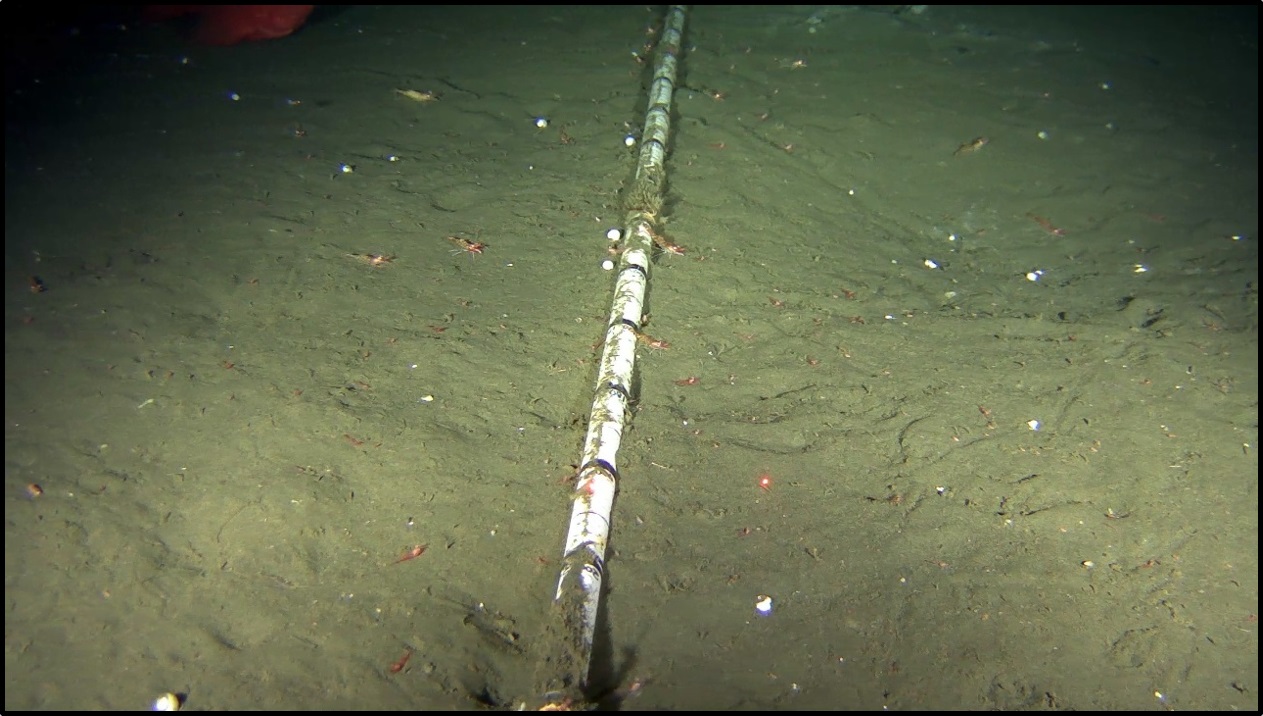}}\vspace{-0.4em}
\subfigure[]
{\includegraphics[width=0.46\linewidth]{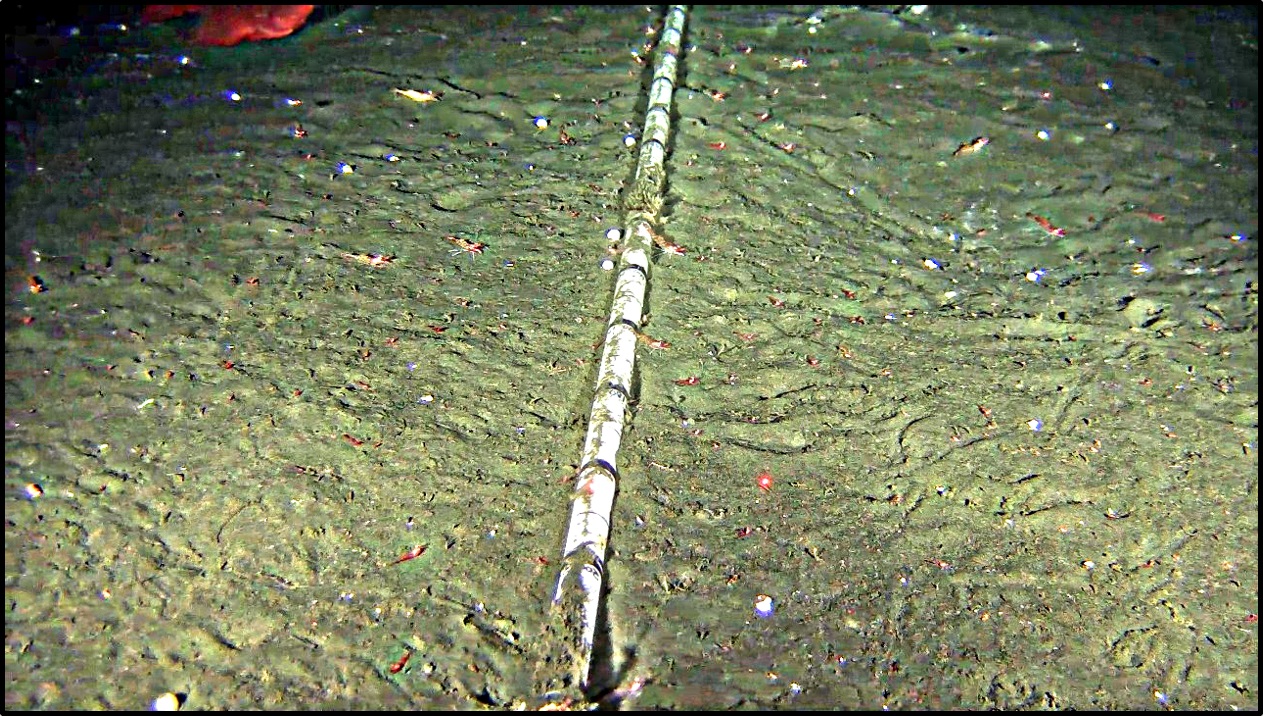}}\vspace{-0.4em}
\subfigure[]
{\includegraphics[width=0.46\linewidth]{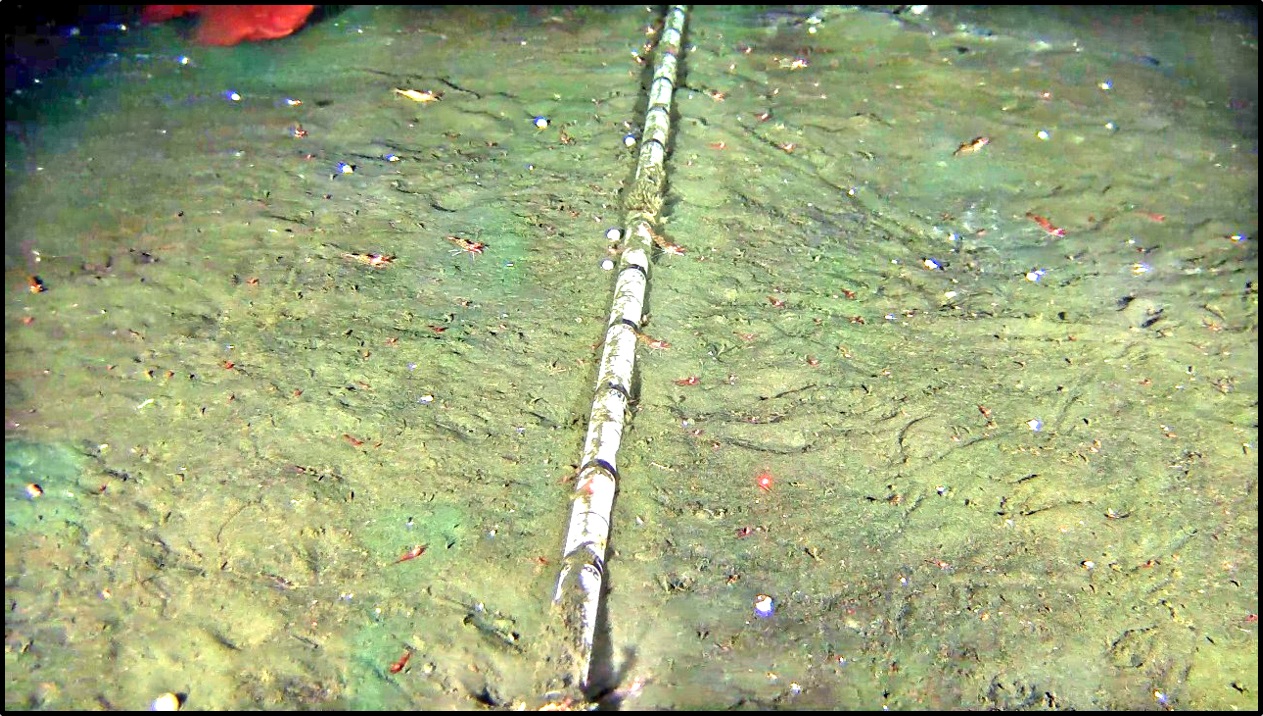}}
\subfigure[]
{\includegraphics[width=0.46\linewidth]{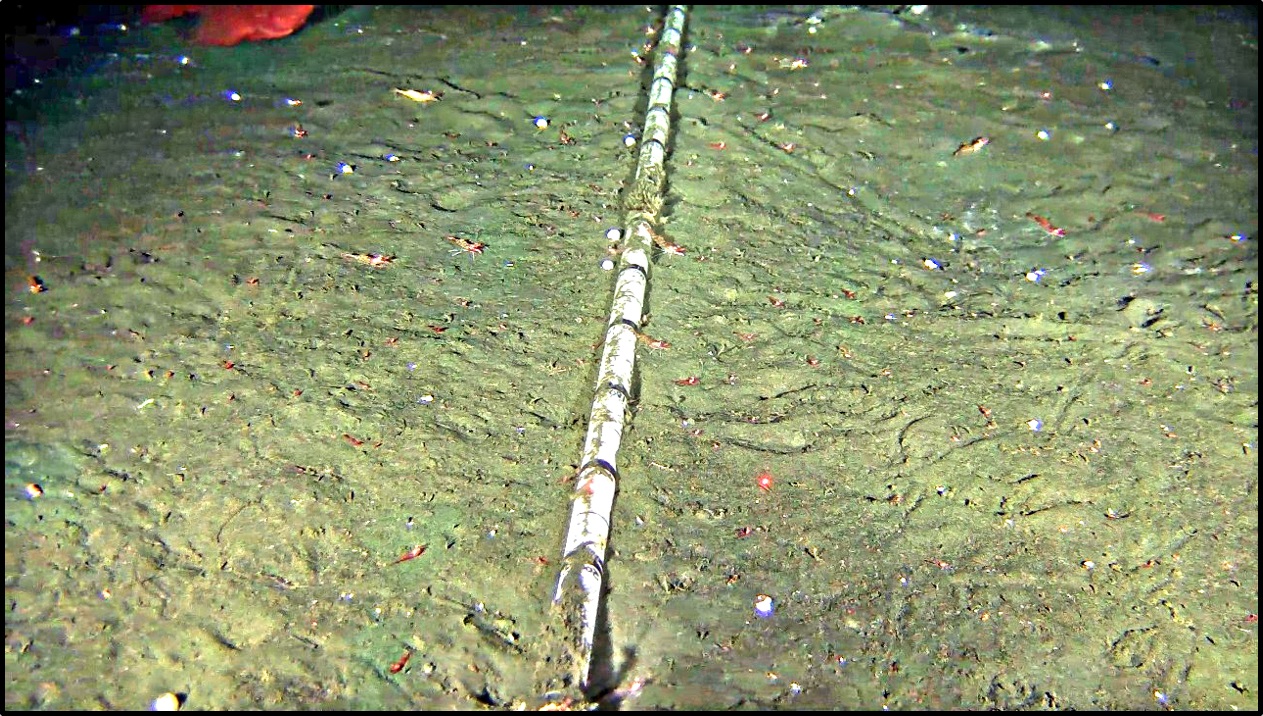}\label{fig:inputs_output_3}}
   \vspace{-1.1em}
   \caption{ \textbf{(a)} Original sample from OceanDark. \textbf{(b)} Input calculated with $m=5$. \textbf{(c)} Input calculated with $m=30$. \textbf{(d)} Final output of \luwe,~a multi-scale fusion between the two inputs.}
\label{fig:inputs_output}
\end{figure}

 Following the procedure of \cite{ancuti2016night,ancuti2013single,ancuti2017color,burt1983laplacian}, each input $\mathcal{I}^{k}$ is decomposed into a $5$-level Laplacian pyramid $L\{\mathcal{I}^{k}\}$. The multi-scale fusion process is then carried out using Equation~\ref{eq:multi_scale_fusion}. The output of \luwe~(Figure \ref{fig:inputs_output_3}) reduces the darkness from the original image, retains colors and enhances important visual features.
 
\section{Experimental results}

The performance of \luwe~is evaluated using the OceanDark dataset \cite{mdpipaper}, composed by 183 samples of low-light underwater images captured by Ocean Networks Canada (ONC). Seven metrics are used in a comprehensive comparison between \luwe~and five underwater-specific image enhancers, as well as two low-light-specific image enhancers. \par

\textbf{Metrics}. Evaluating the results of image enhancers in datasets without ground truth (such as OceanDark~\cite{mdpipaper}) is non-trivial. Individual metrics can not, alone, indicate the performance of the enhancement; e.g., the $e$-score~\cite{hautiere2011blind} measures the number of new visible edges obtained after enhancement, which can represent noise, while the FADE score~\cite{choi2015referenceless} can be used to measure darkness, but it does not account for visual features lost because of over-illumination. Thus we employed a group of seven metrics in the evaluation. \textbf{UIQM}~\cite{panetta2015human}: inspired on the human visual system, this no-reference underwater image quality indicator combines measures of colorfulness, sharpness and contrast; \textbf{PCQI}~\cite{wang2015patch}: a method developed to assess the quality of contrast-changed images, PCQI considers local contrast quality maps; \textbf{GCF}~\cite{matkovic2005global}: this metric reflects the level of contrast present in the whole image (dehazed images typically present higher contrast); \textbf{e-} and \textbf{r-scores}~\cite{hautiere2011blind}: indicate, based on the original and enhanced images, the increase in number of visible edges, and the boost in gradient value for these edge's pixels (``visibility''), respectively; \textbf{FADE}~\cite{choi2015referenceless}: measures the amount of perceived fog (or darkness in inverted images), thus assigning, in our analysis, lower values to better-illuminated images; \textbf{SURF}~\cite{bay2008speeded}: a popular method that extracts useful features for image matching, reconstruction, stitching, object detection, among others.\par

\textbf{Comparison with state-of-the-art methods}. We evaluate \luwe~against five frameworks designed specifically for the enhancement of underwater images: Marques \etal~\cite{mdpipaper}, Berman \etal~\cite{berman2017diving}, Drews \etal~\cite{drews2013transmission}, Fu \etal~\cite{fu2014retinex} and Cho \etal~\cite{cho2017visibility}. Since OceanDark~\cite{mdpipaper} is composed of low-light images, we also considered two popular low-light-specific image enhancers: Zhang \etal~\cite{zhang2017fast} and Guo \etal~\cite{guo2016lime}. All methods are discussed in sub-section~\ref{ss:background}. The implementations used are those made publicly available by the authors.\par

\textbf{Qualitative analysis}. Figure~\ref{fig:experiment} shows that \luwe~is able to generate output images that highlights important visual features (e.g., fishes that were not visible, small rocks and overall geography of the sites) of the original images without excessively brightening the  scenes. While the methods of Marques \etal~\cite{mdpipaper} and Guo~\etal~\cite{guo2016lime} yielded similar results that greatly reduced low-light regions, they also concealed important visual features because of a lighting over-correction: close-to-saturation pixel intensities might hide the image's finer details, such as edges and textures. The methods of Drews~\etal~\cite{drews2013transmission}, Zhang~\etal~\cite{zhang2017fast} and Cho~\etal~\cite{cho2017visibility} actually darkened the images, contrary to the goal of the enhancement. The methods of Zhang~\etal~\cite{zhang2017fast} and Fu~\etal~\cite{fu2014retinex} highlighted edges and textures of the outputs, but were not able to properly illuminate dark regions. The unnatural colors generated by the method of Berman~\etal~\cite{berman2017diving} can be attributed to the automatic, but often sub-optimal, choice of water type done for each sample. \par

\textbf{Quantitative analysis}. \luwe~obtained the highest score on the UIQM~\cite{panetta2015human} metric, confirming the human-visual-system-inspired perceived image quality of its outputs. The quality of the contrast modification measured by PCQI~\cite{wang2015patch} also favors the results obtained with the proposed method. The framework of Zhang~\etal~\cite{zhang2017fast} obtained the highest GCF~\cite{matkovic2005global}, however a qualitative analysis of its results indicates that this is due to the strong saturation of the output, which creates a high \textit{global} contrast with respect to the undesired dark regions still present in the enhanced images. The highest increase in visible edges ($e$-score~\cite{hautiere2011blind}) and SURF features~\cite{bay2008speeded} is attributed to \luwe~(we reiterate that these two metrics should be interpreted cautiously, as they may reflect noise introduced to the image). Finally, the highest scores in the metrics related to visibility ($r$-score~\cite{hautiere2011blind}) and FADE~\cite{choi2015referenceless} indicate that \luwe~successfully achieved the goal of increasing the illumination of low-light underwater images while preserving their colors and highlighting their salient visual structures.

\section{Conclusion}

Our novel single-image framework, \luwe, uses a contrast-guided approach for the efficient modelling of lighting distributions in low-light underwater scenes. It then generates two dehazed inputs that are combined employing a multi-scale fusion process, ultimately reducing dark regions and highlighting important visual features of the original image without changing its color distribution.\par
\begin{figure*}
\begin{center}
\footnotesize
\begin{minipage}[c]{0.095\textwidth}
\textbf{Original} \label{fig:experiment_original}
\end{minipage}%
\begin{minipage}[c]{0.9\linewidth}
\includegraphics[width=0.195\textwidth]{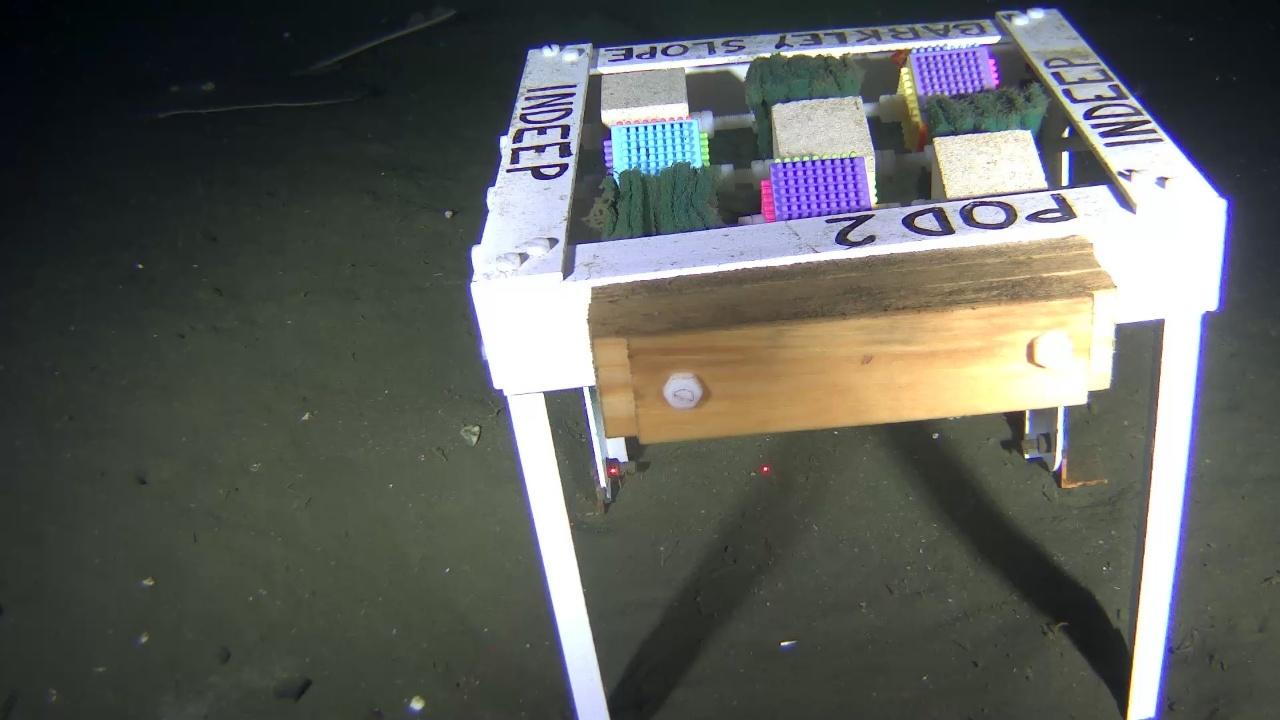}
\includegraphics[width=0.195\textwidth]{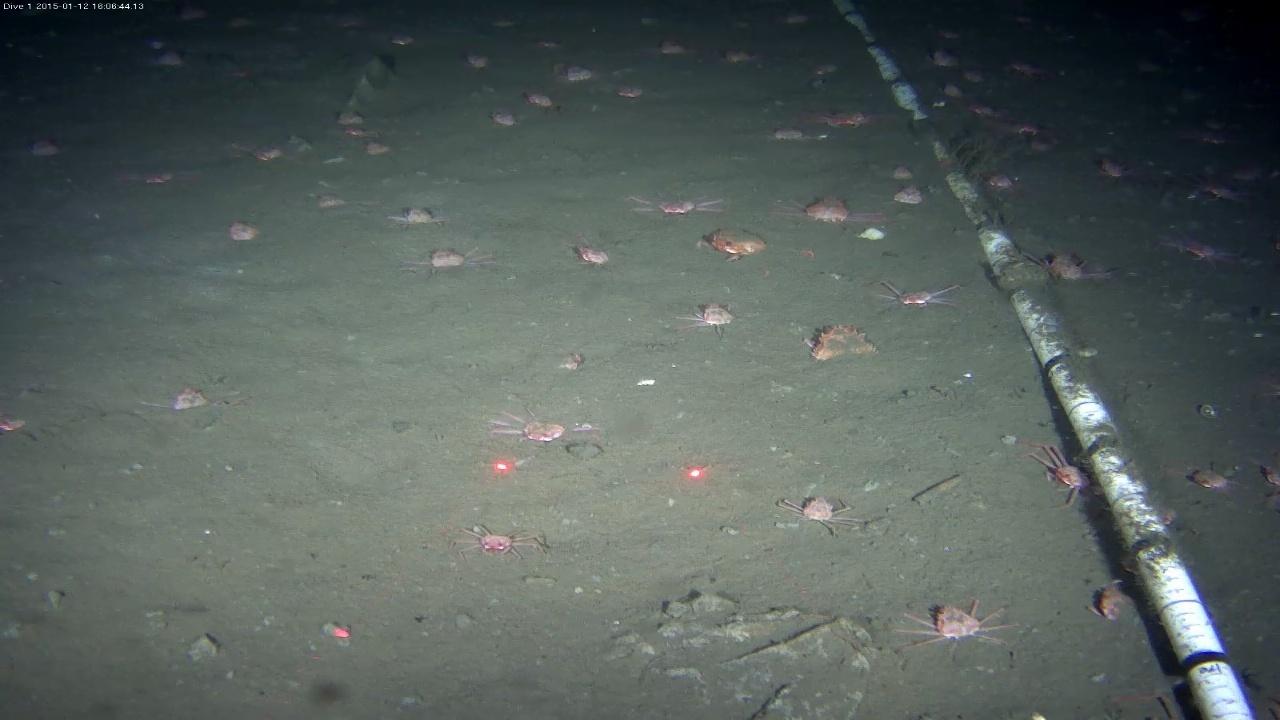}
\includegraphics[width=0.195\textwidth]{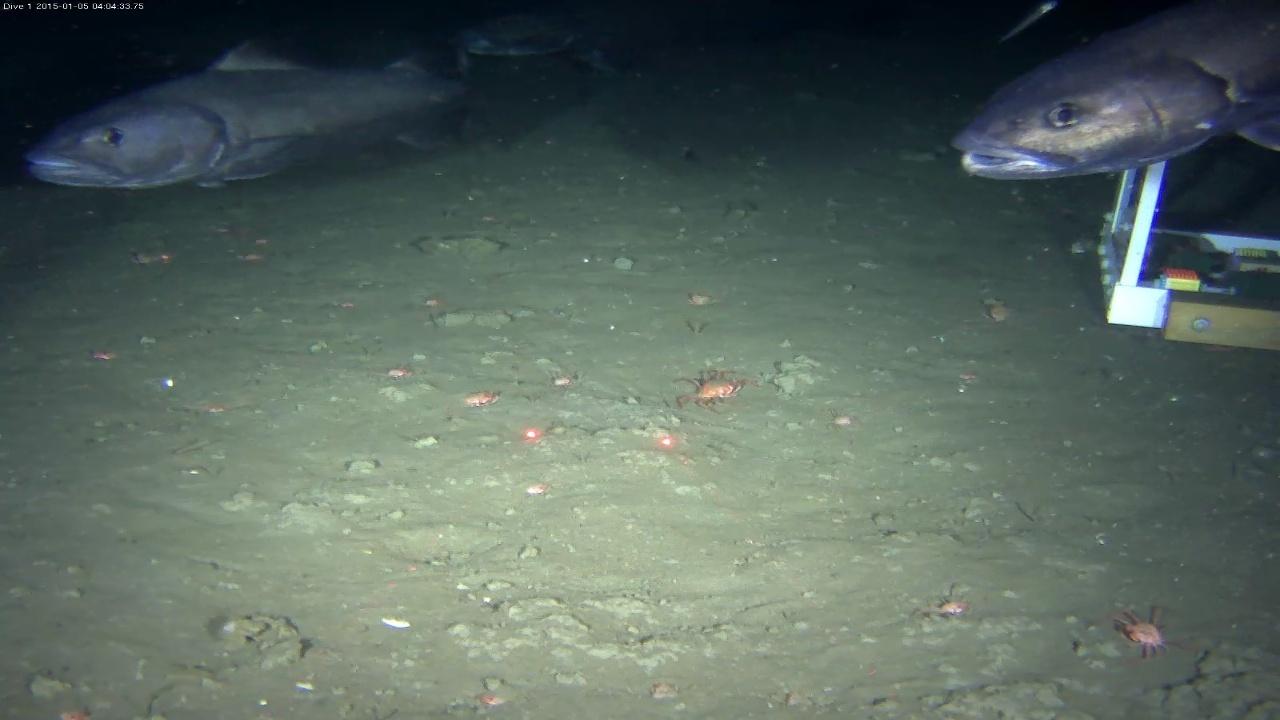}
\includegraphics[width=0.195\textwidth]{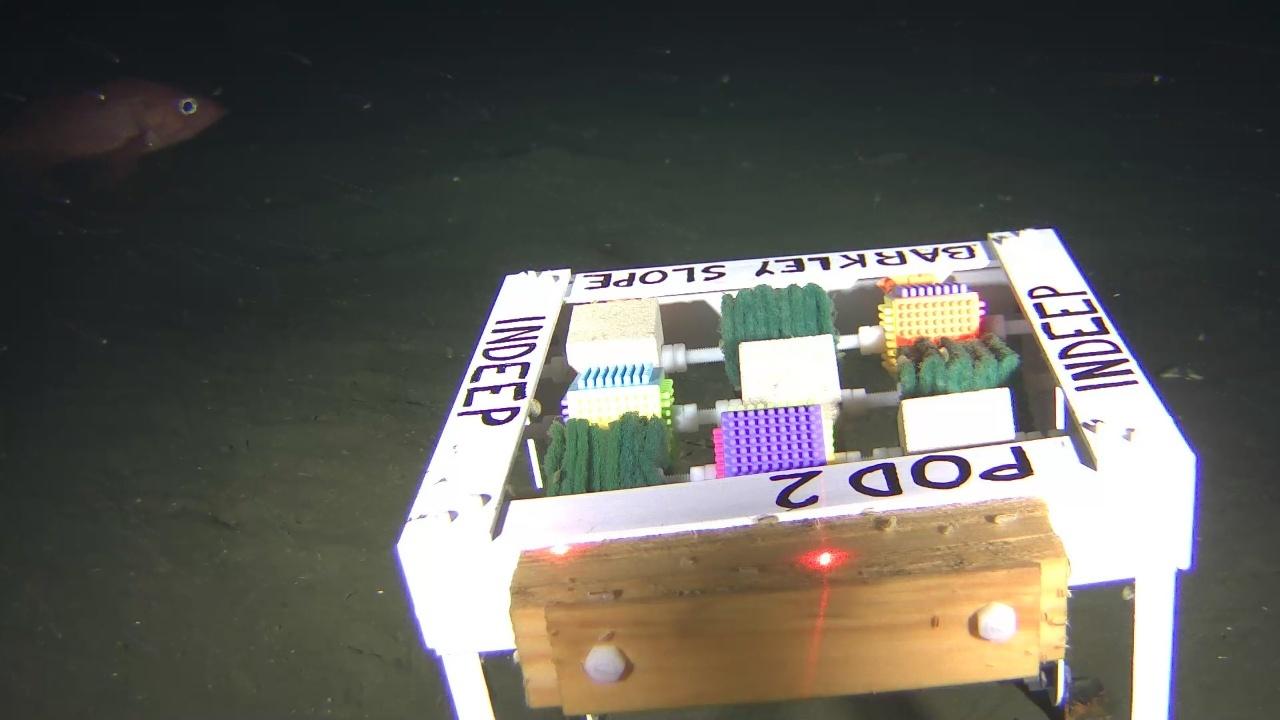}
\includegraphics[width=0.195\textwidth]{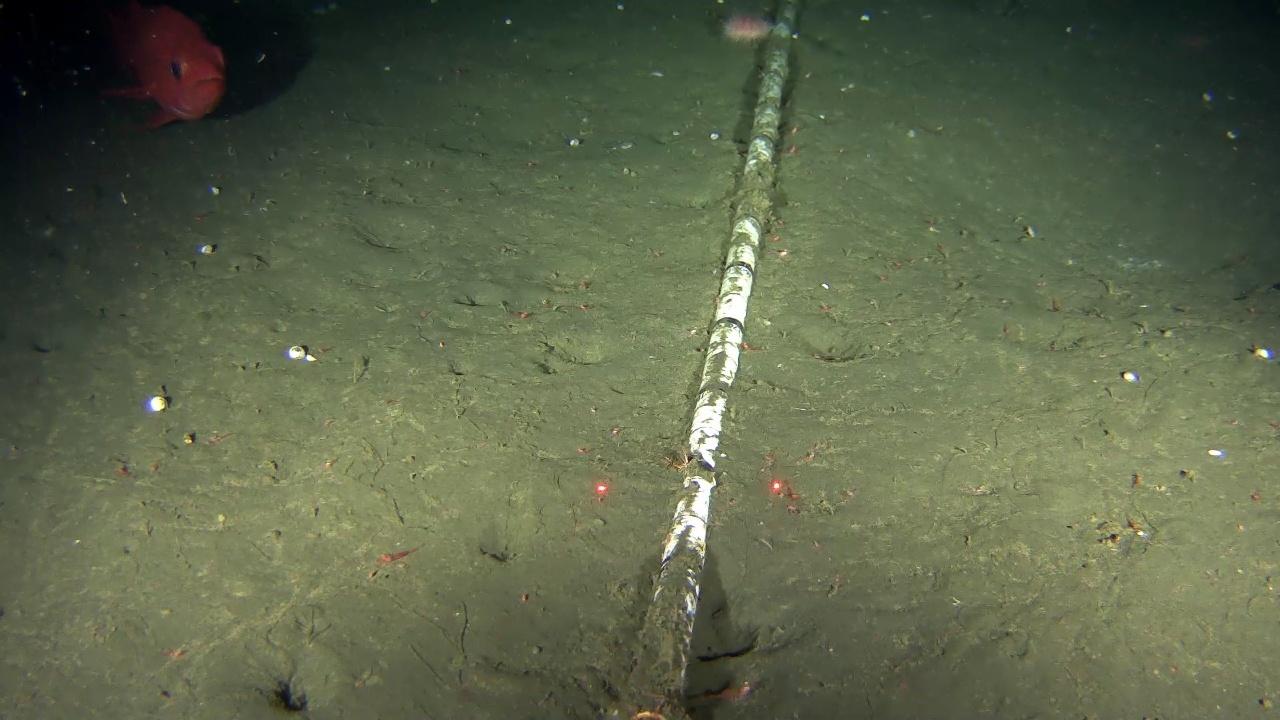}
\end{minipage}\hfill%

\begin{minipage}[c]{0.095\textwidth}
\textbf{\luwe} \label{fig:experiment_luwe}
\end{minipage}%
\begin{minipage}[c]{0.9\linewidth}
\includegraphics[width=0.195\textwidth]{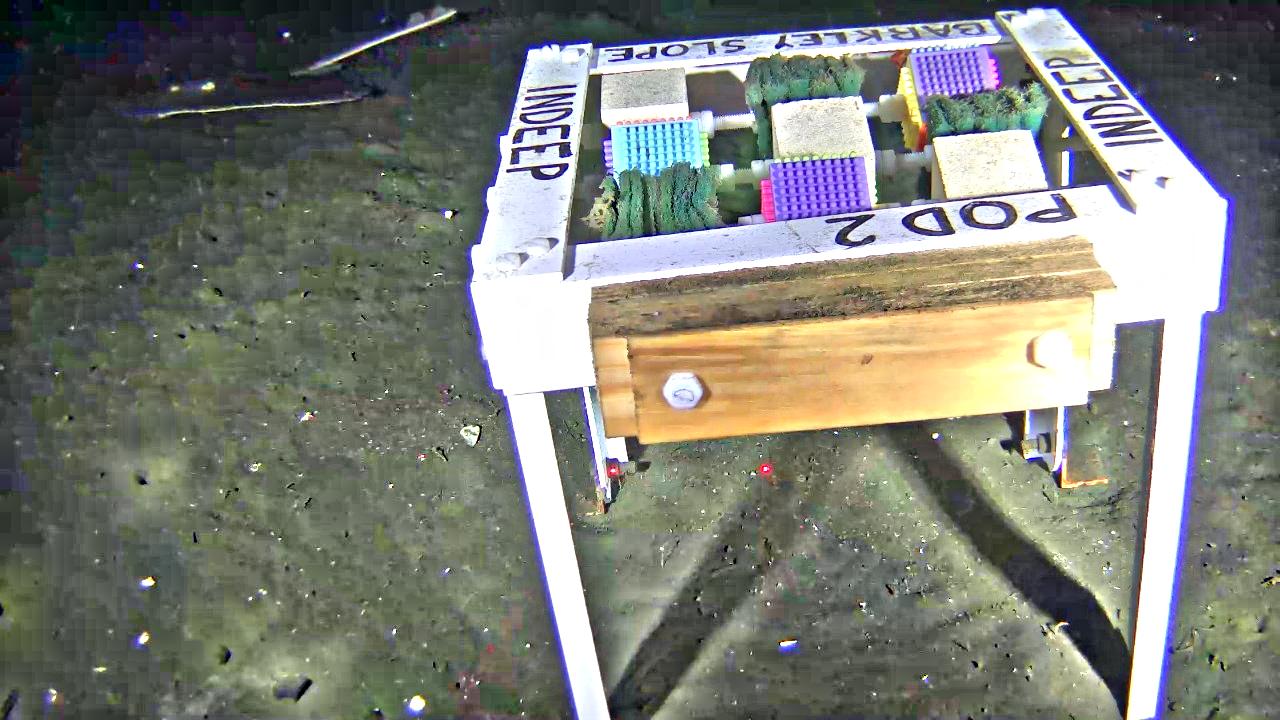}
\includegraphics[width=0.195\textwidth]{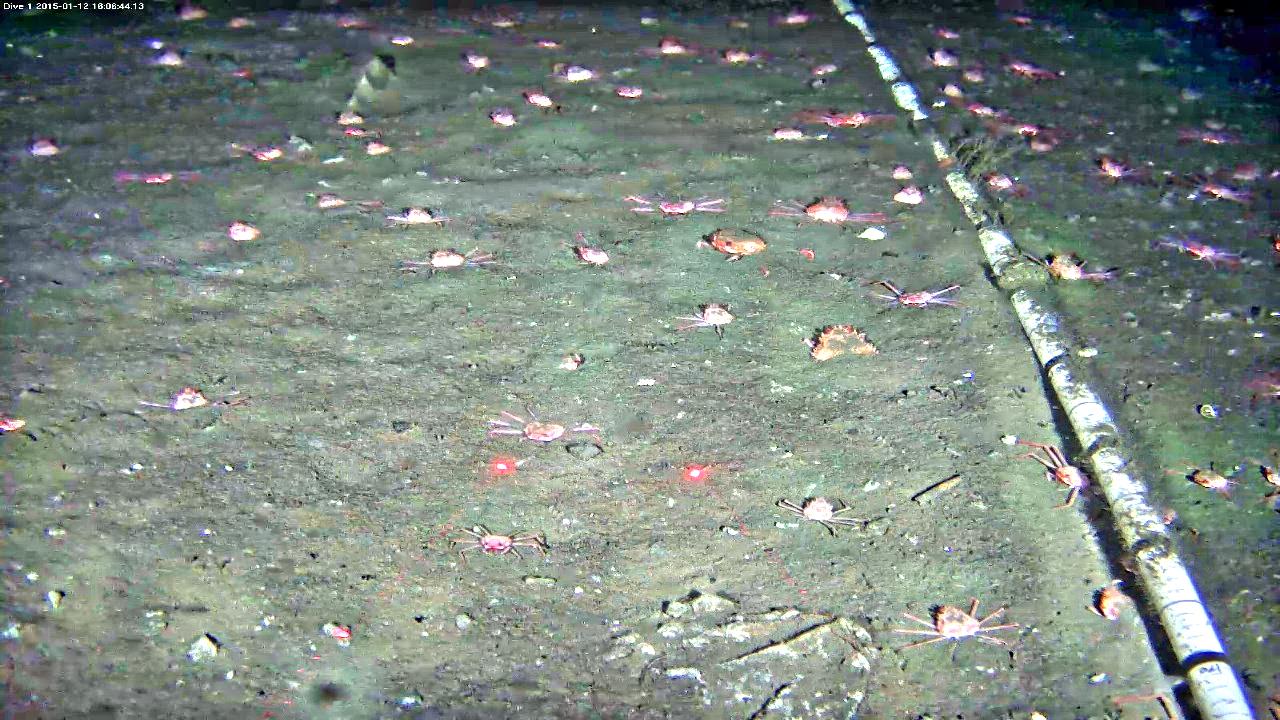}
\includegraphics[width=0.195\textwidth]{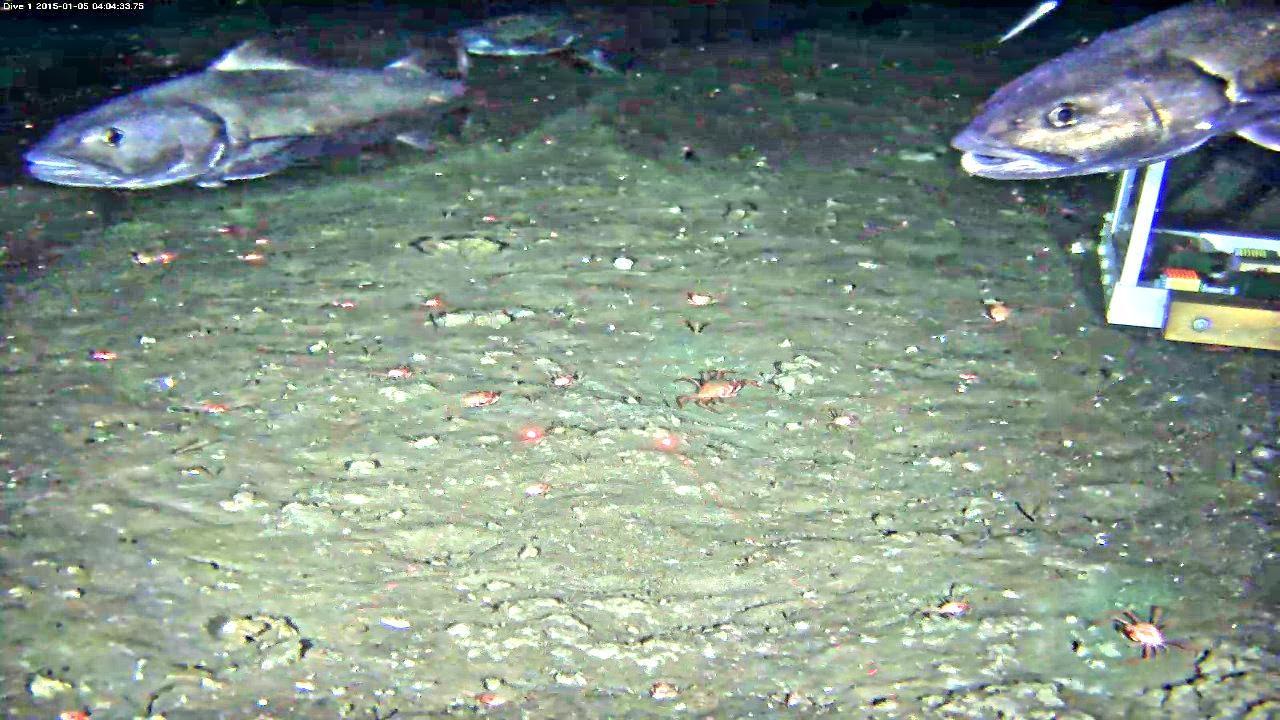}
\includegraphics[width=0.195\textwidth]{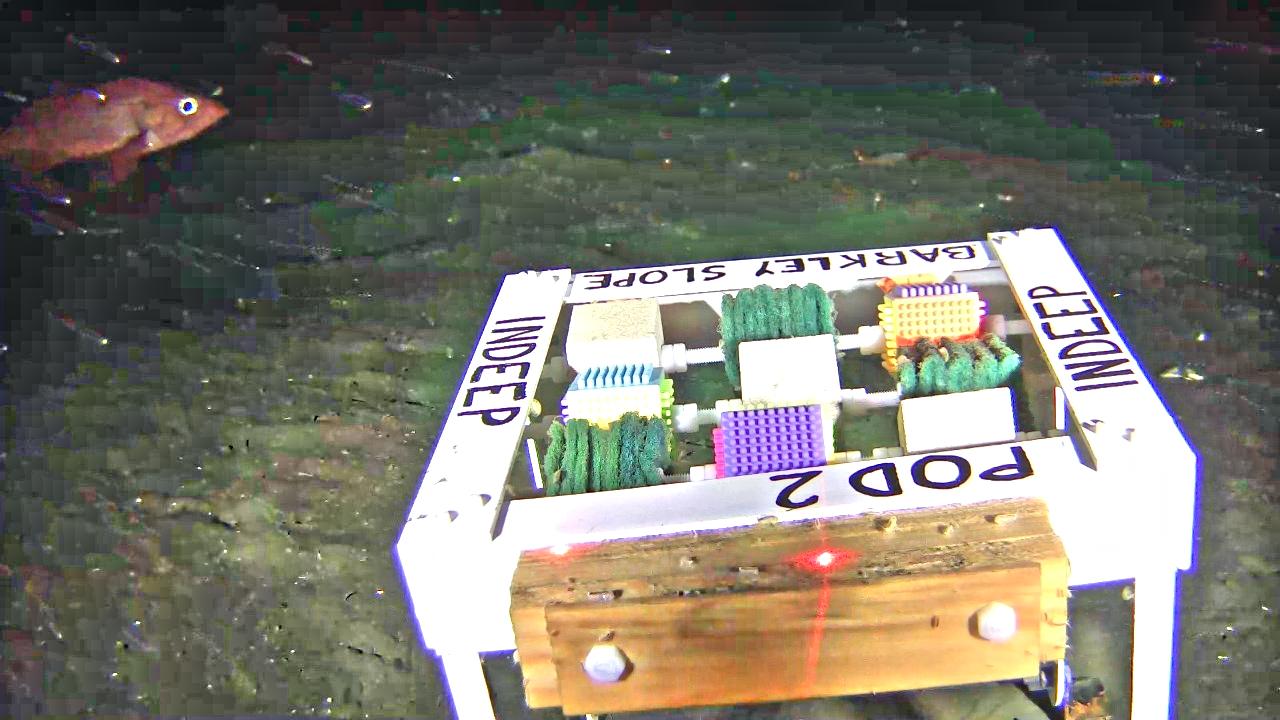}
\includegraphics[width=0.195\textwidth]{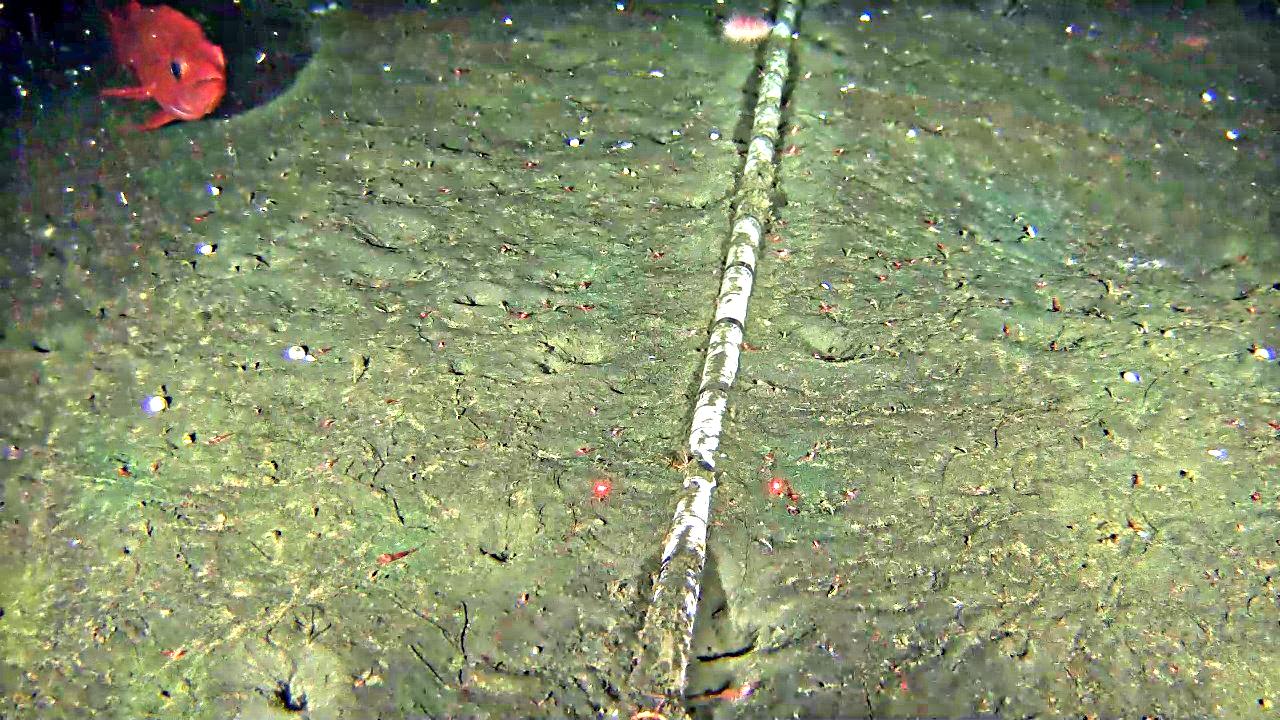}
\end{minipage}\hfill

\begin{minipage}[c]{0.095\textwidth}
\textbf{Marques}~\cite{mdpipaper} \label{fig:experiment_marques}
\end{minipage}%
\begin{minipage}[c]{0.9\linewidth}
\includegraphics[width=0.195\textwidth]{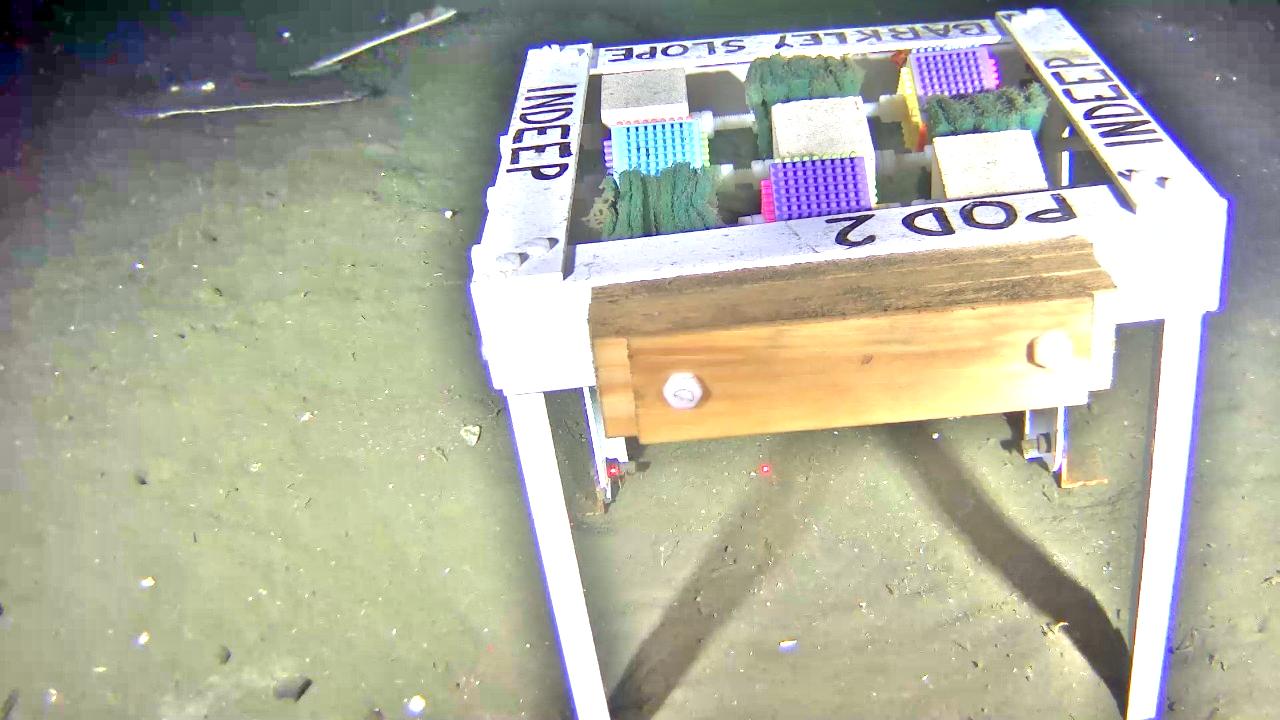}
\includegraphics[width=0.195\textwidth]{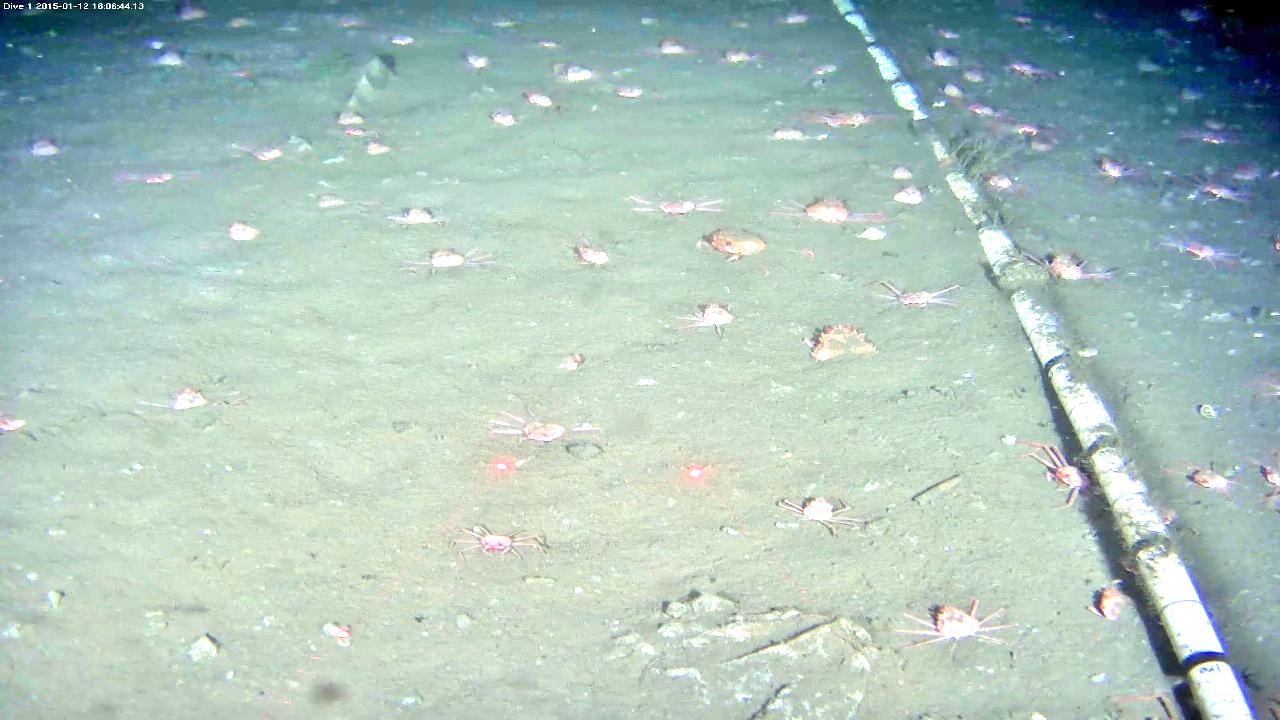}
\includegraphics[width=0.195\textwidth]{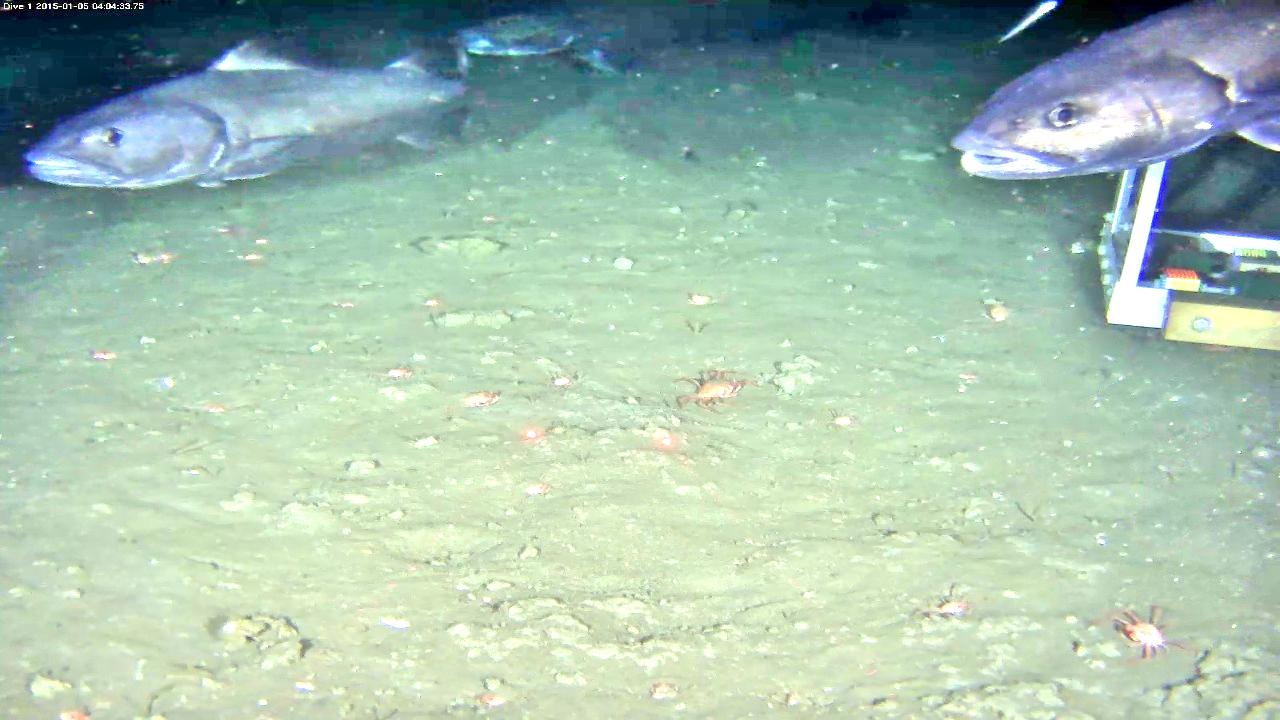}
\includegraphics[width=0.195\textwidth]{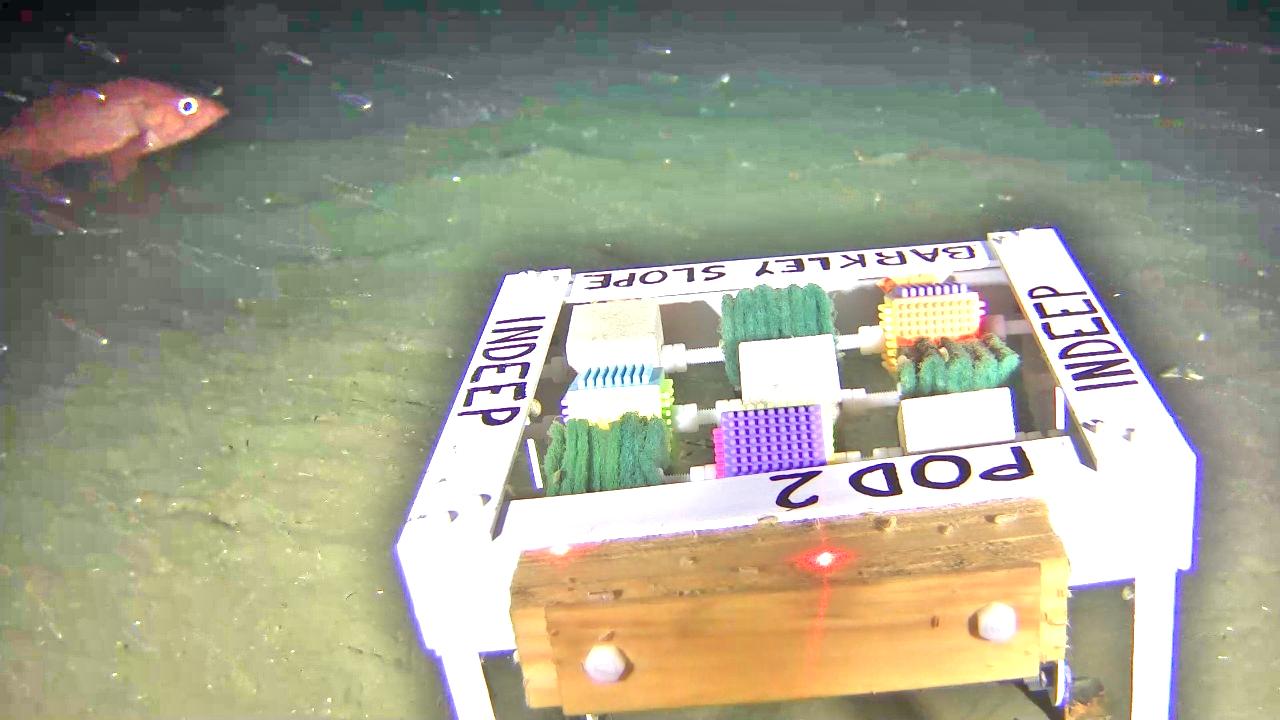}
\includegraphics[width=0.195\textwidth]{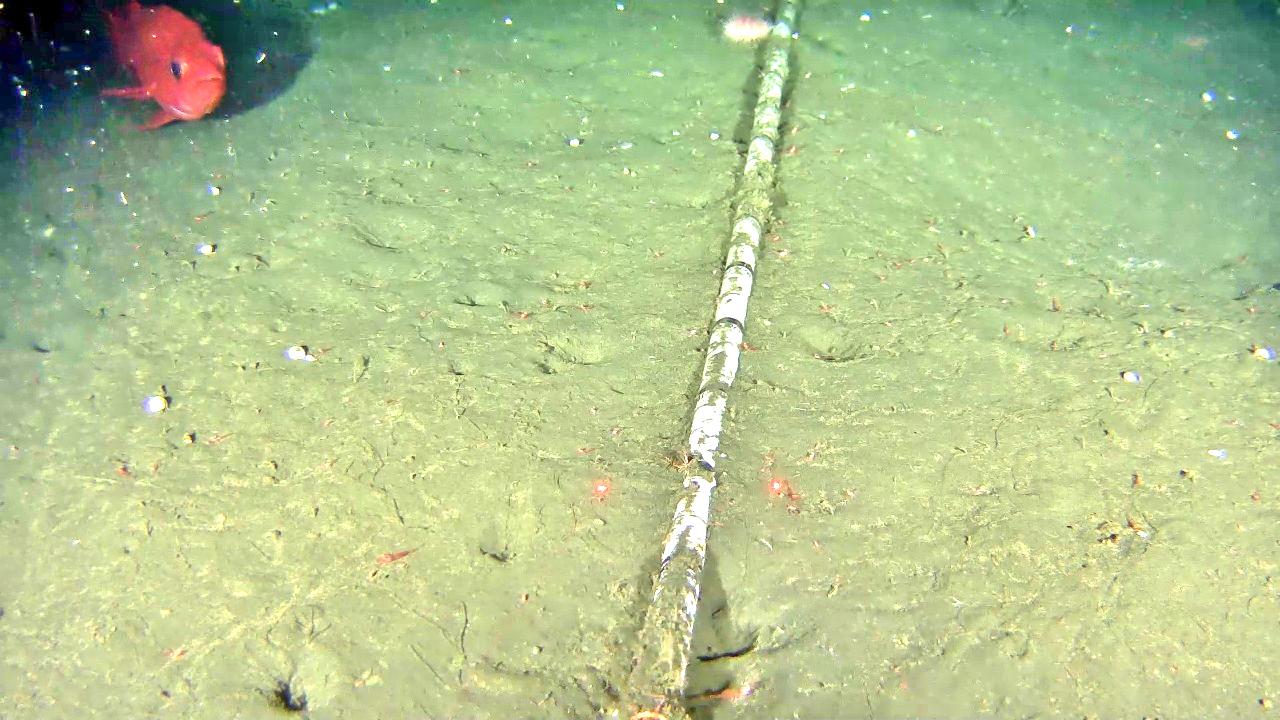}
\end{minipage}\hfill

\begin{minipage}[c]{0.095\textwidth}
\textbf{Berman}~\cite{berman2017diving} \label{fig:experiment_berman}
\end{minipage}%
\begin{minipage}[c]{0.9\linewidth}
\includegraphics[width=0.195\textwidth]{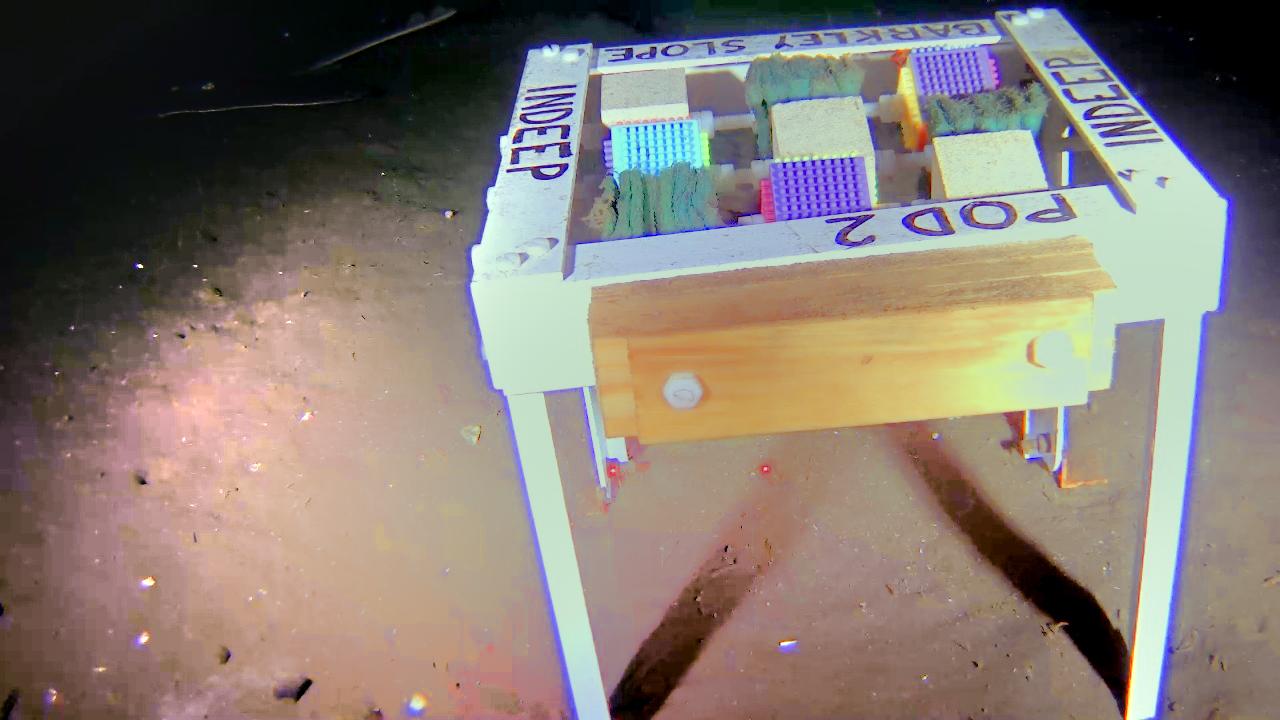}
\includegraphics[width=0.195\textwidth]{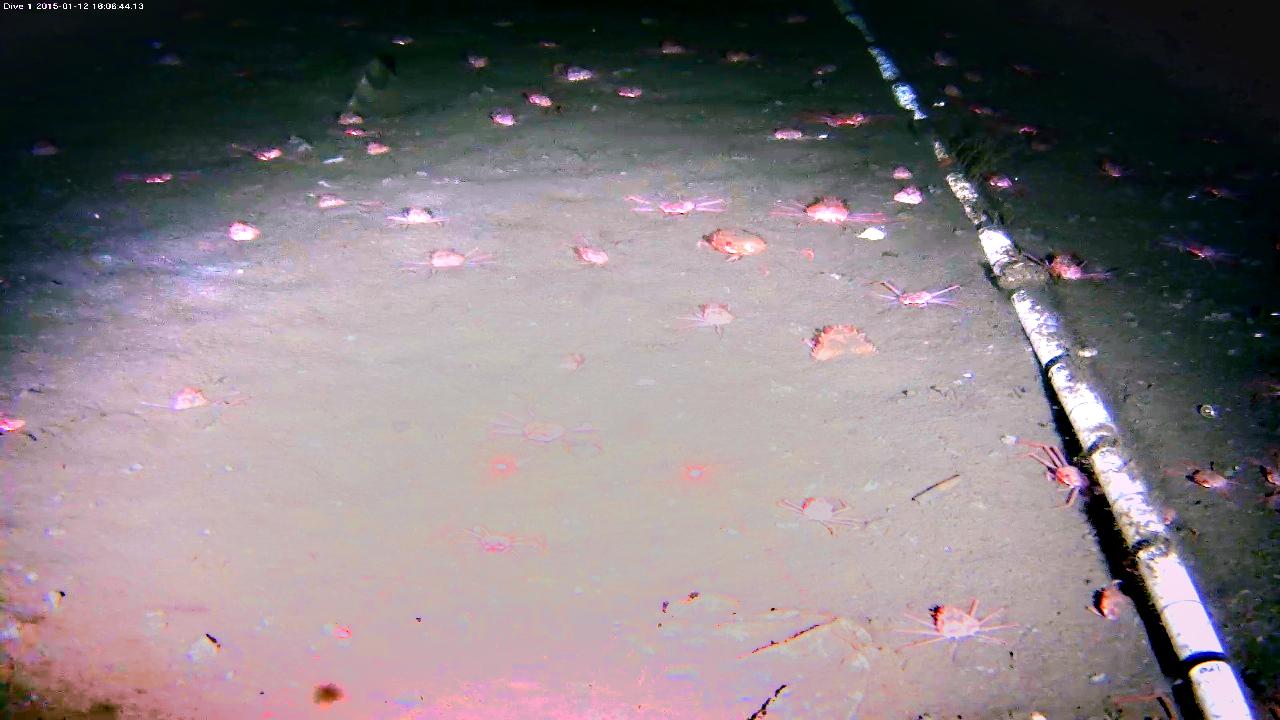}
\includegraphics[width=0.195\textwidth]{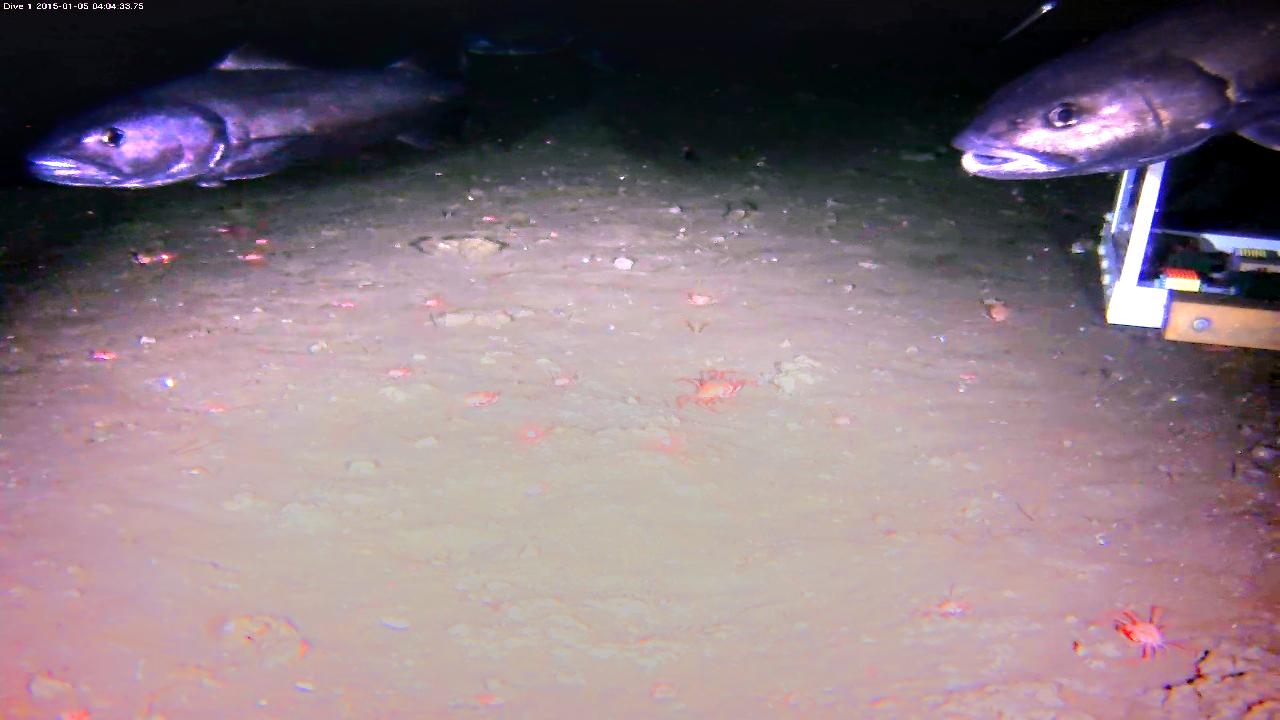}
\includegraphics[width=0.195\textwidth]{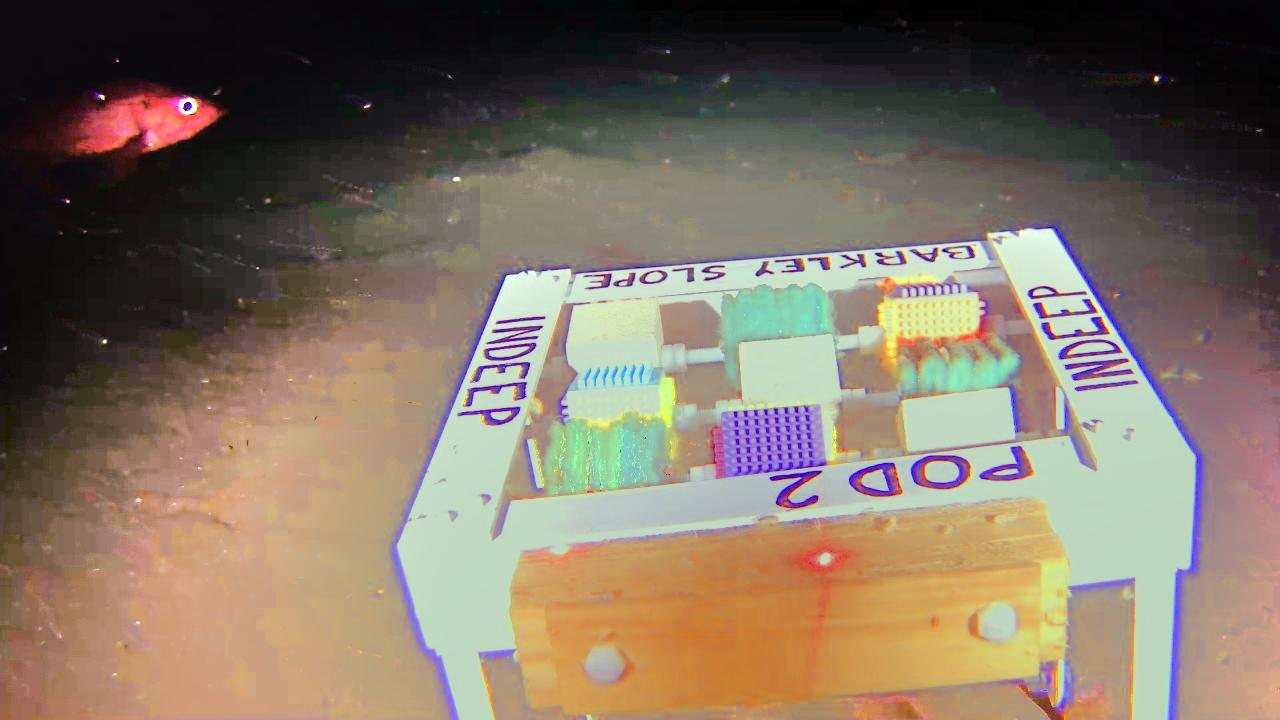}
\includegraphics[width=0.195\textwidth]{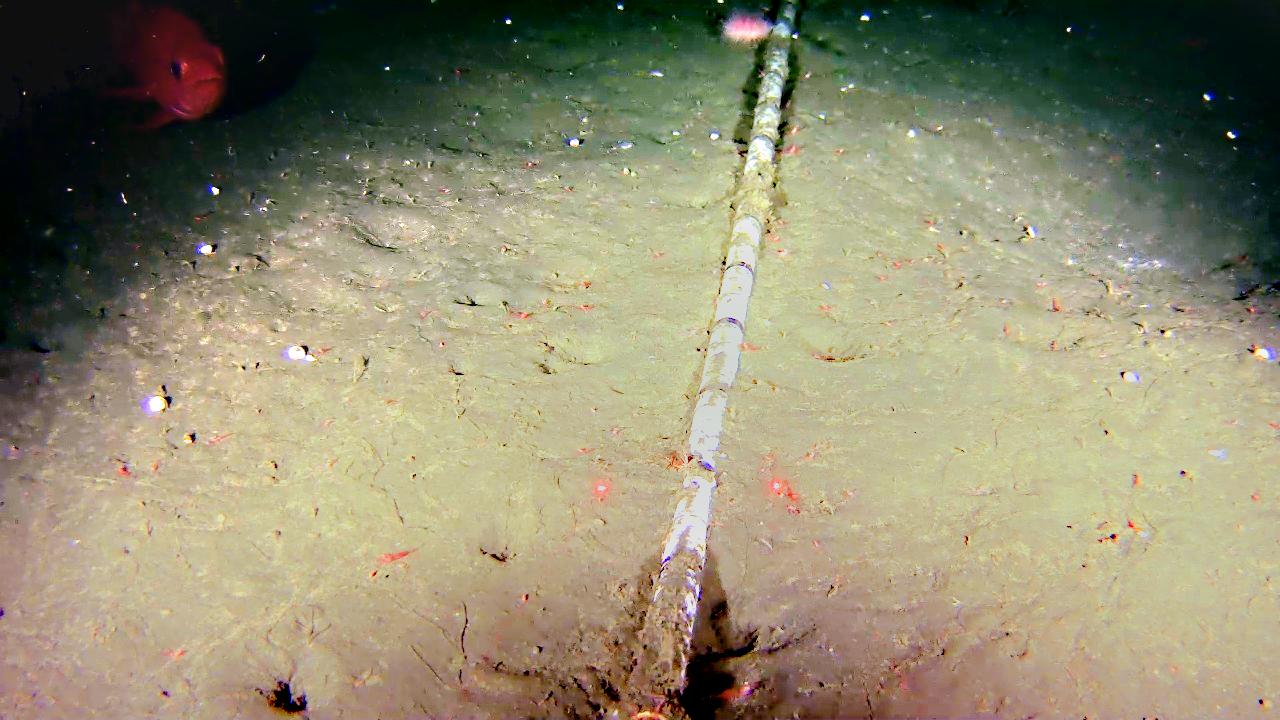}
\end{minipage}\hfill

\begin{minipage}[c]{0.095\textwidth}
\textbf{Fu}~\cite{fu2014retinex} \label{fig:experiment_fu}
\end{minipage}%
\begin{minipage}[c]{0.9\linewidth}
\includegraphics[width=0.195\textwidth]{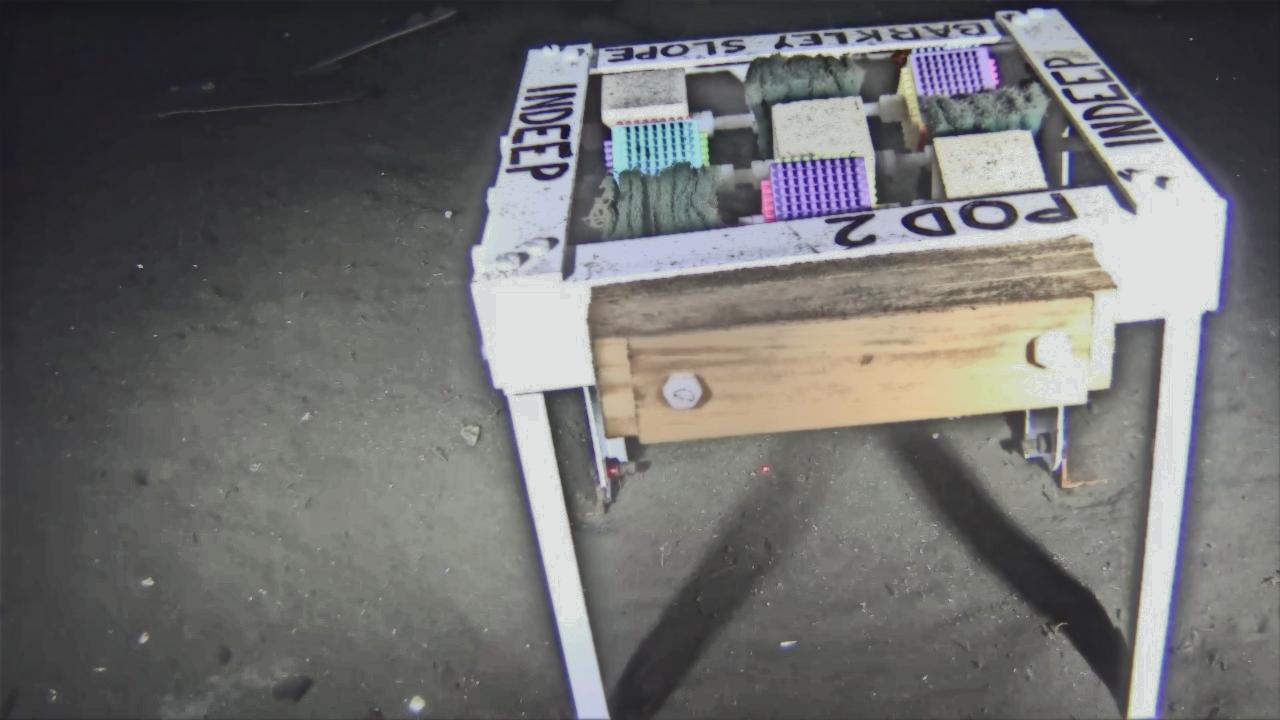}
\includegraphics[width=0.195\textwidth]{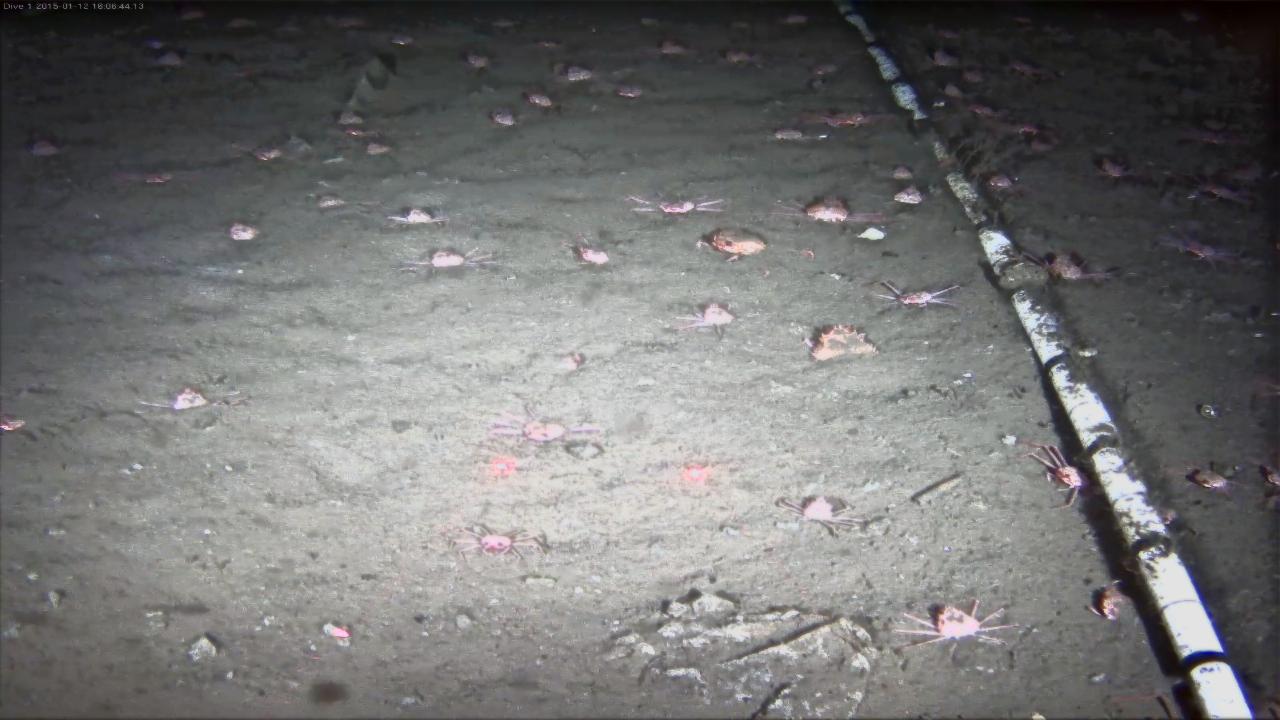}
\includegraphics[width=0.195\textwidth]{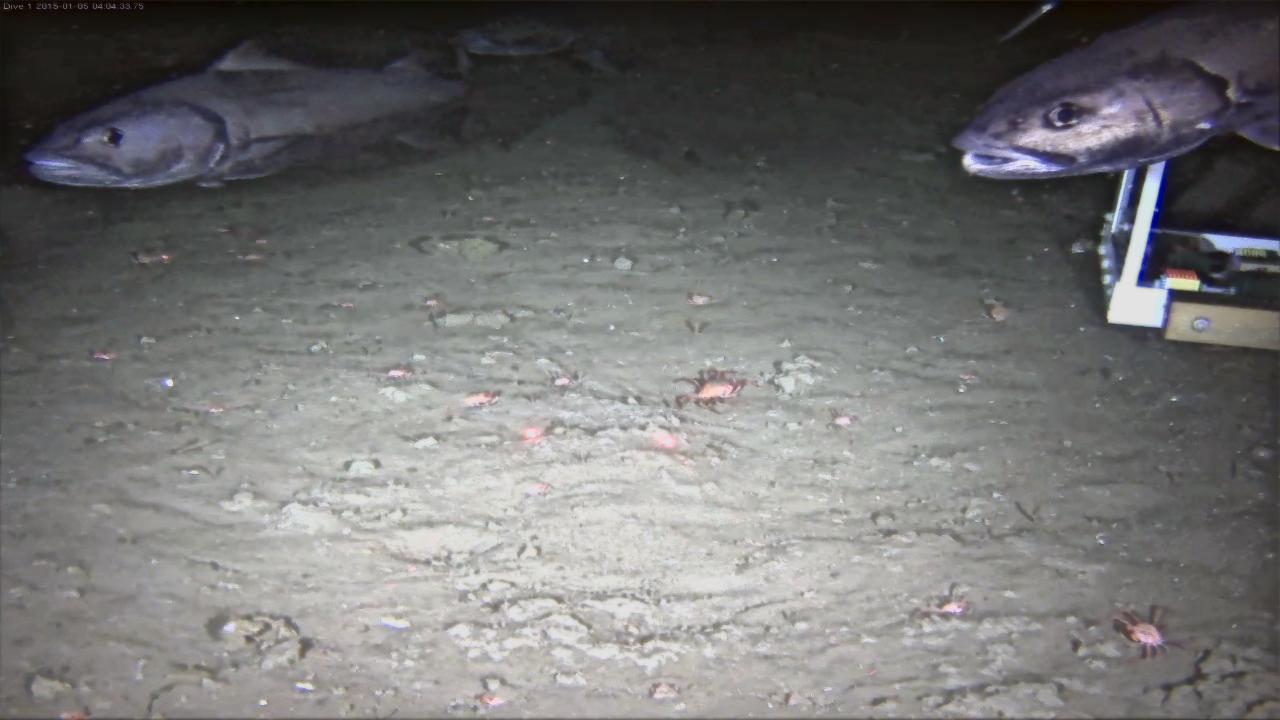}
\includegraphics[width=0.195\textwidth]{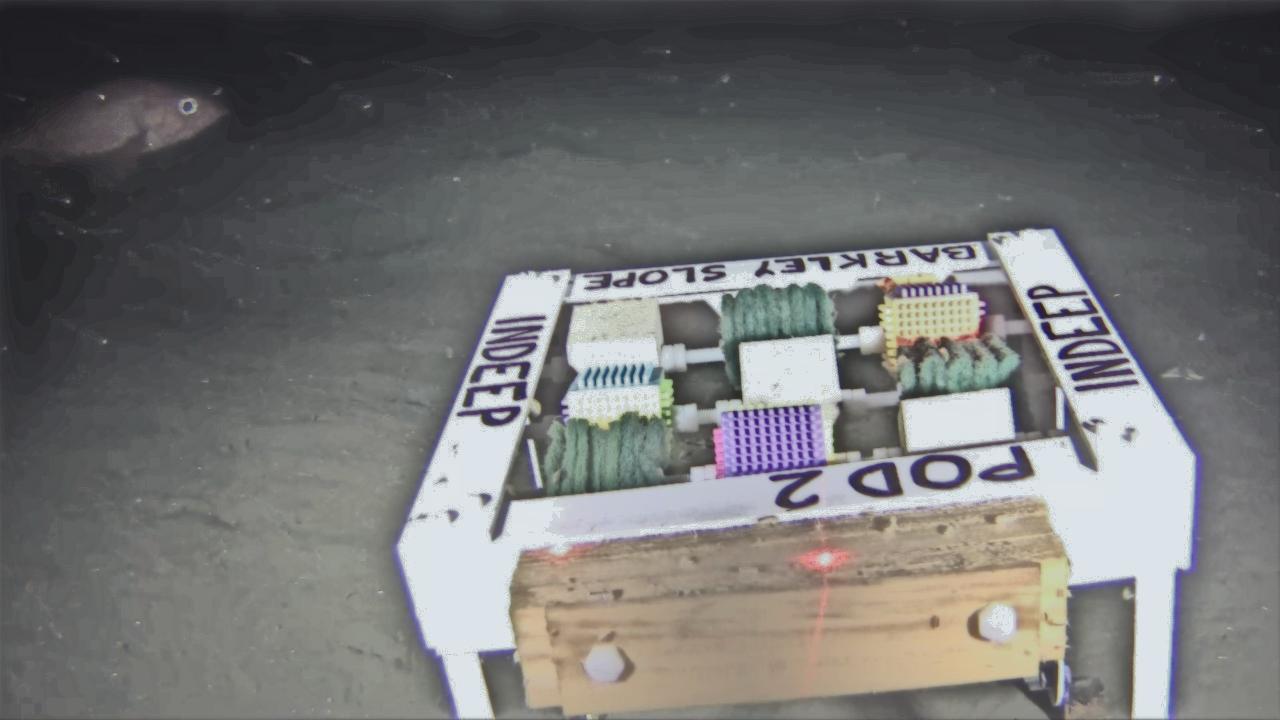}
\includegraphics[width=0.195\textwidth]{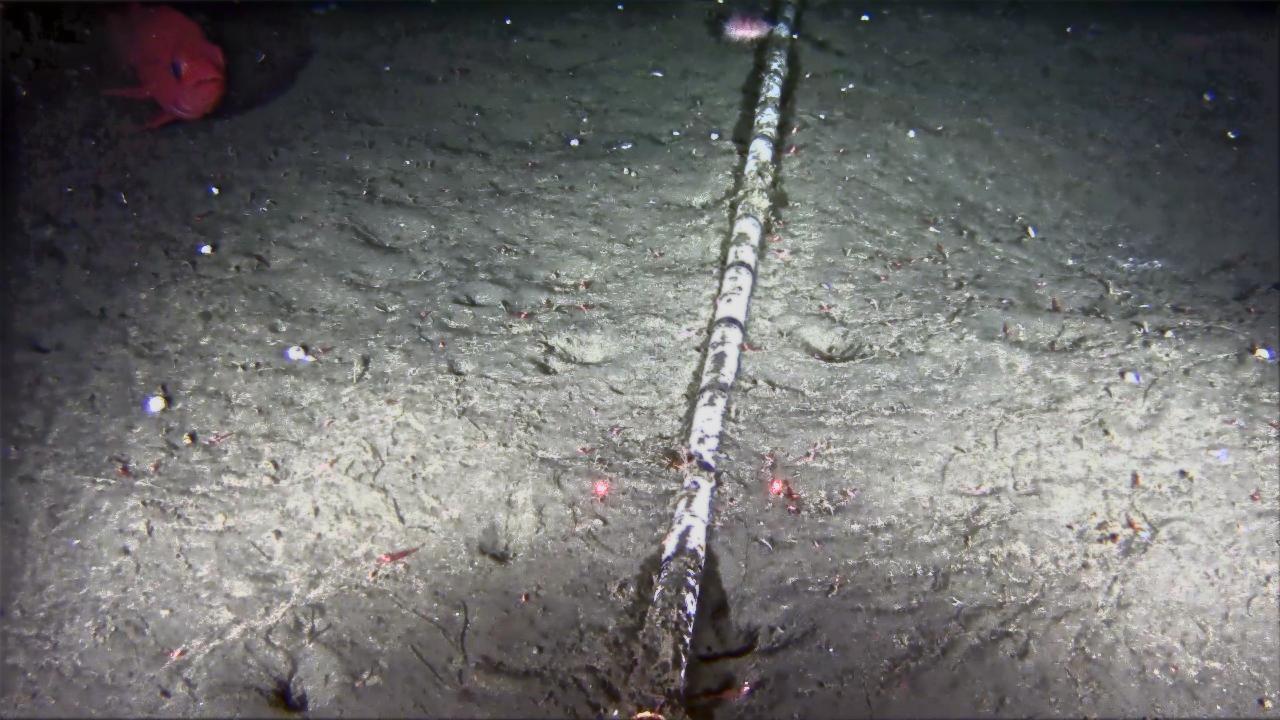}
\end{minipage}\hfill

\begin{minipage}[c]{0.095\textwidth}
\textbf{Cho}~\cite{cho2017visibility} \label{fig:experiment_cho}
\end{minipage}%
\begin{minipage}[c]{0.9\linewidth}
\includegraphics[width=0.195\textwidth]{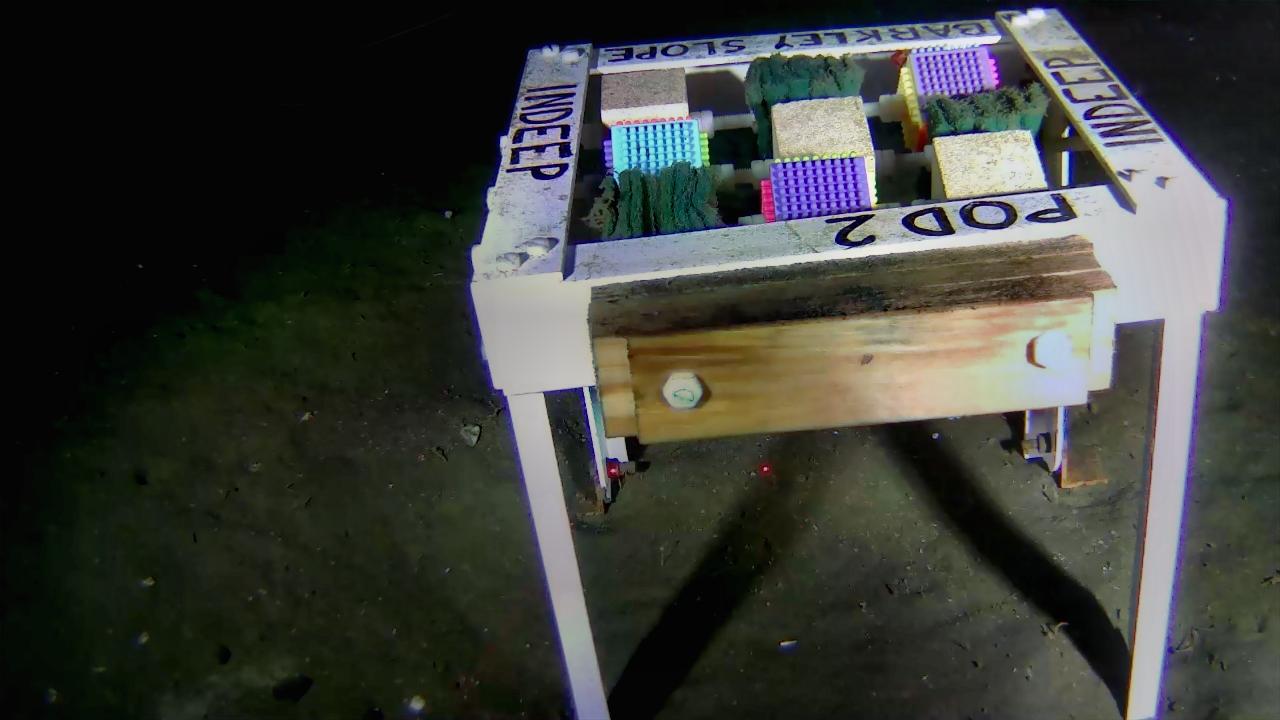}
\includegraphics[width=0.195\textwidth]{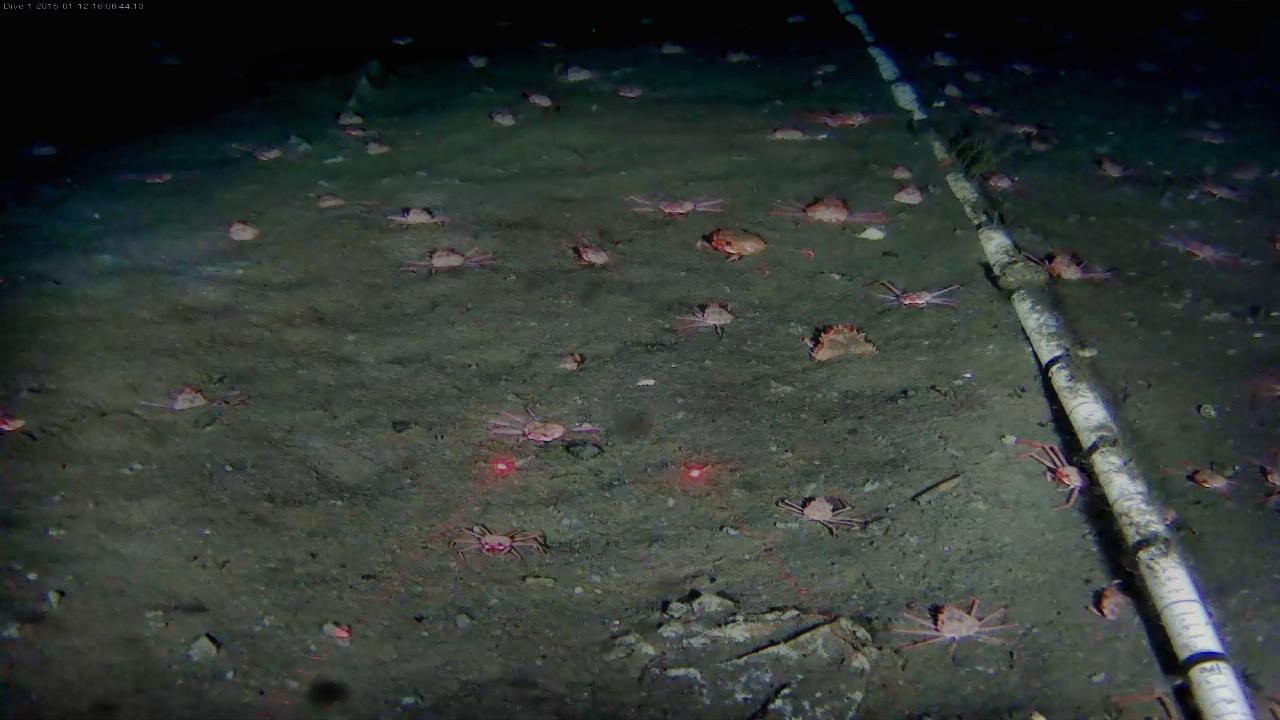}
\includegraphics[width=0.195\textwidth]{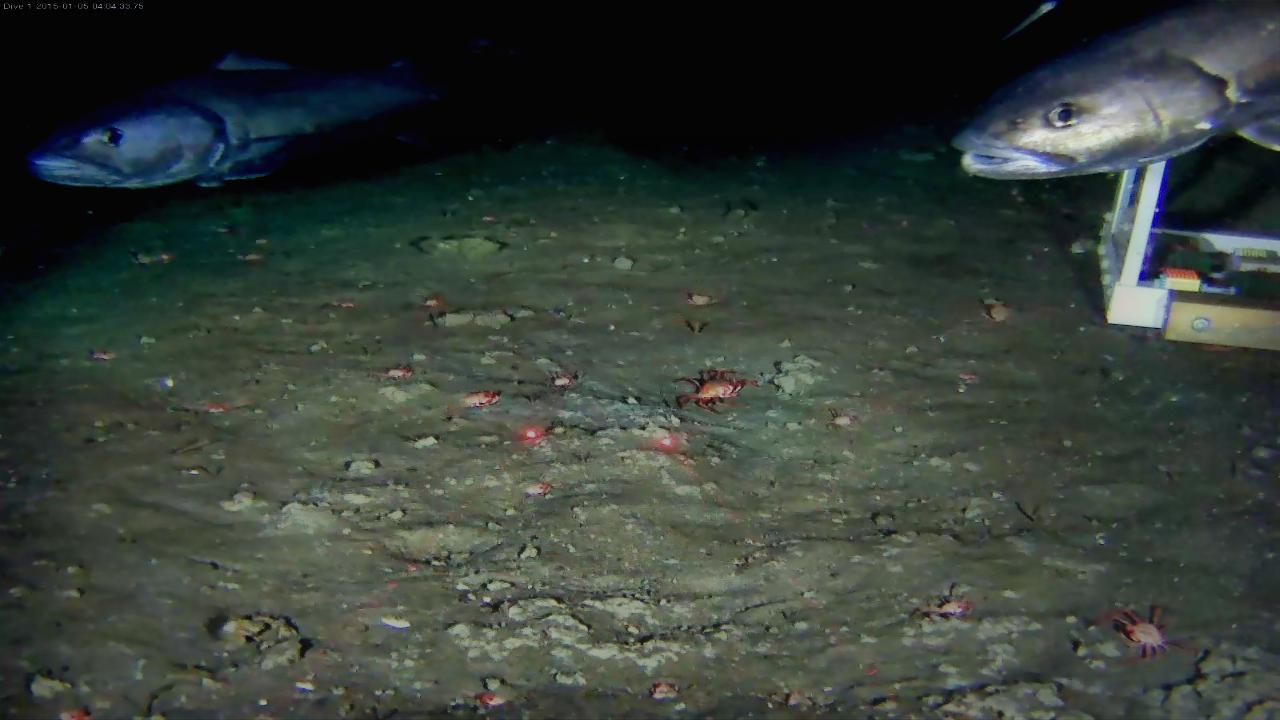}
\includegraphics[width=0.195\textwidth]{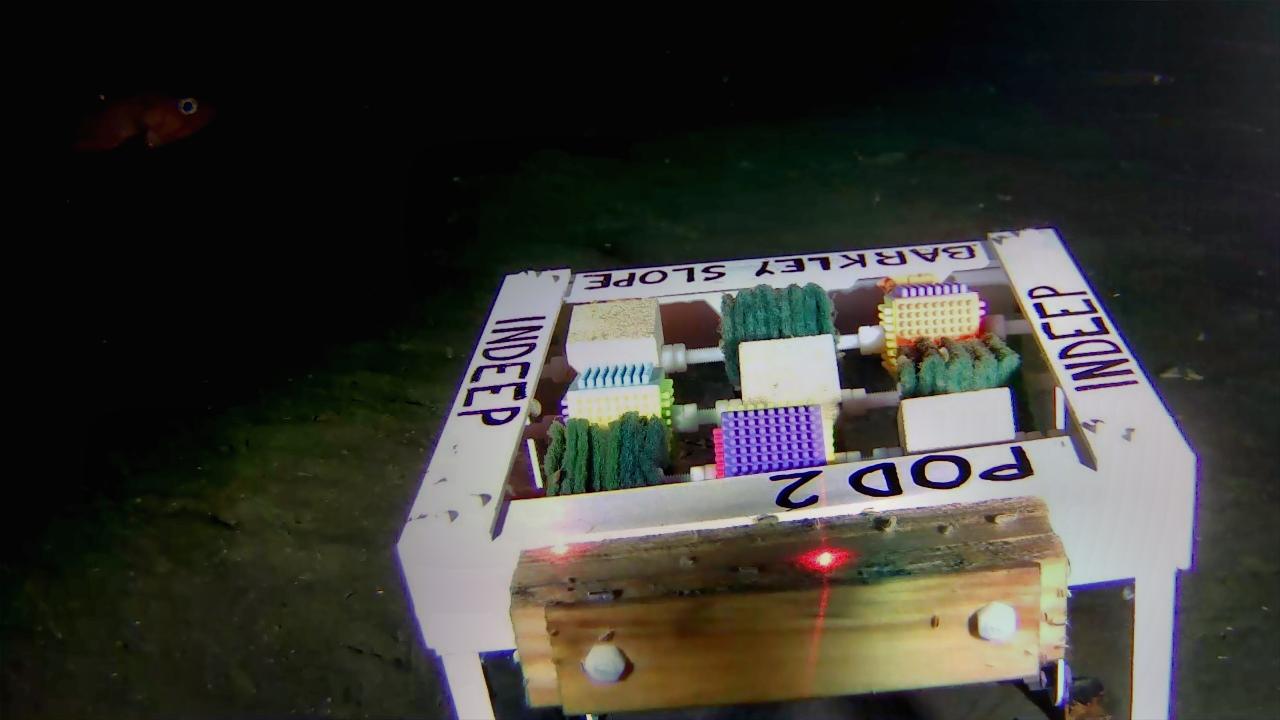}
\includegraphics[width=0.195\textwidth]{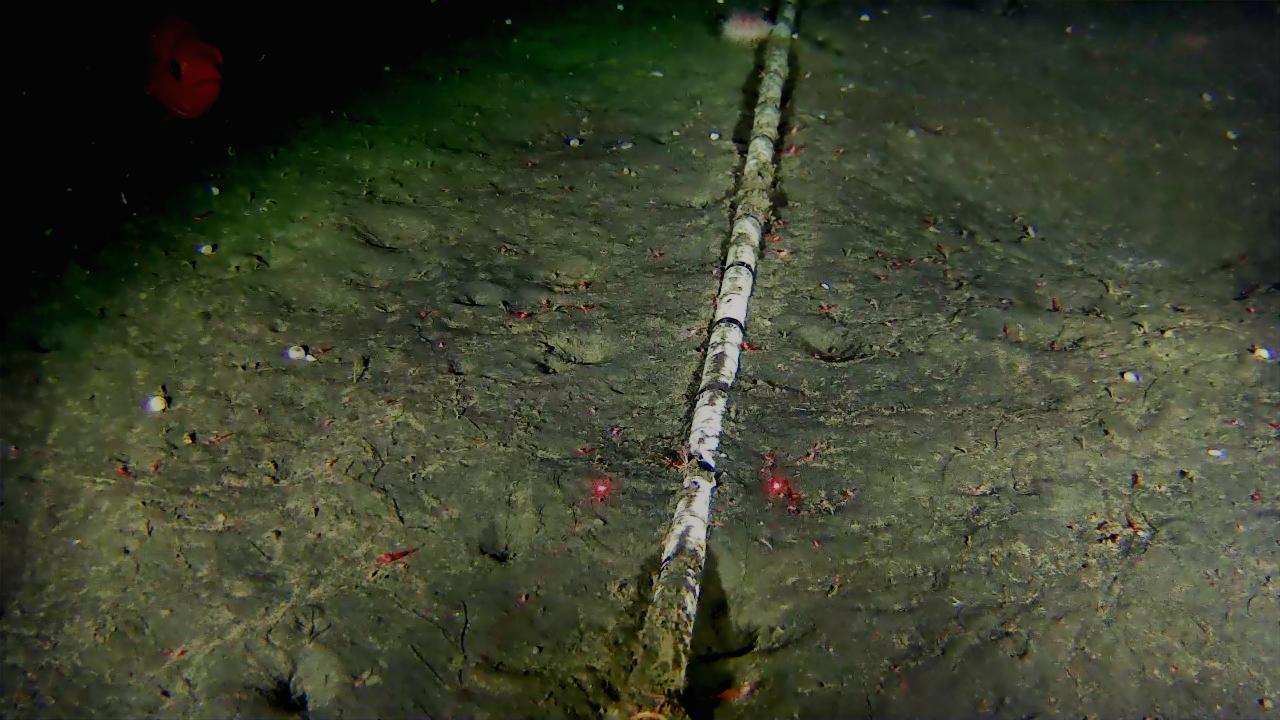}
\end{minipage}\hfill

\begin{minipage}[c]{0.095\textwidth}
\textbf{Drews}~\cite{drews2013transmission} \label{fig:experiment_drews}
\end{minipage}%
\begin{minipage}[c]{0.9\linewidth}
\includegraphics[width=0.195\textwidth]{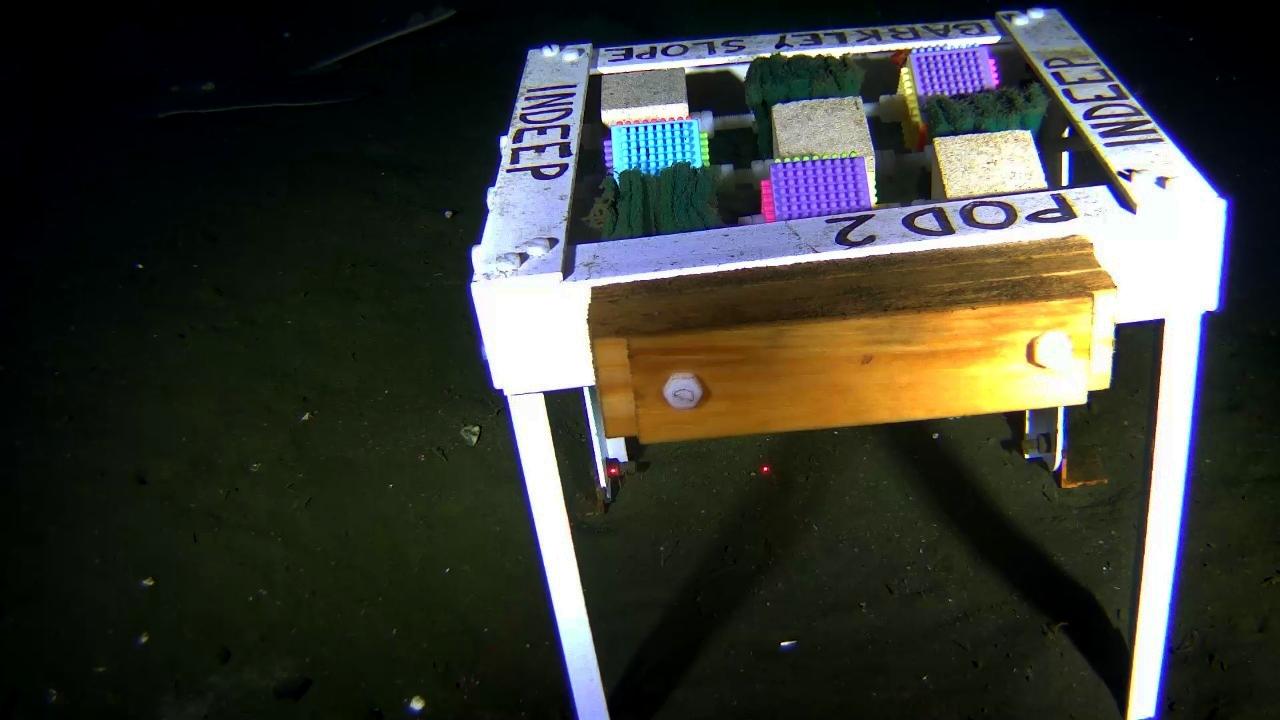}
\includegraphics[width=0.195\textwidth]{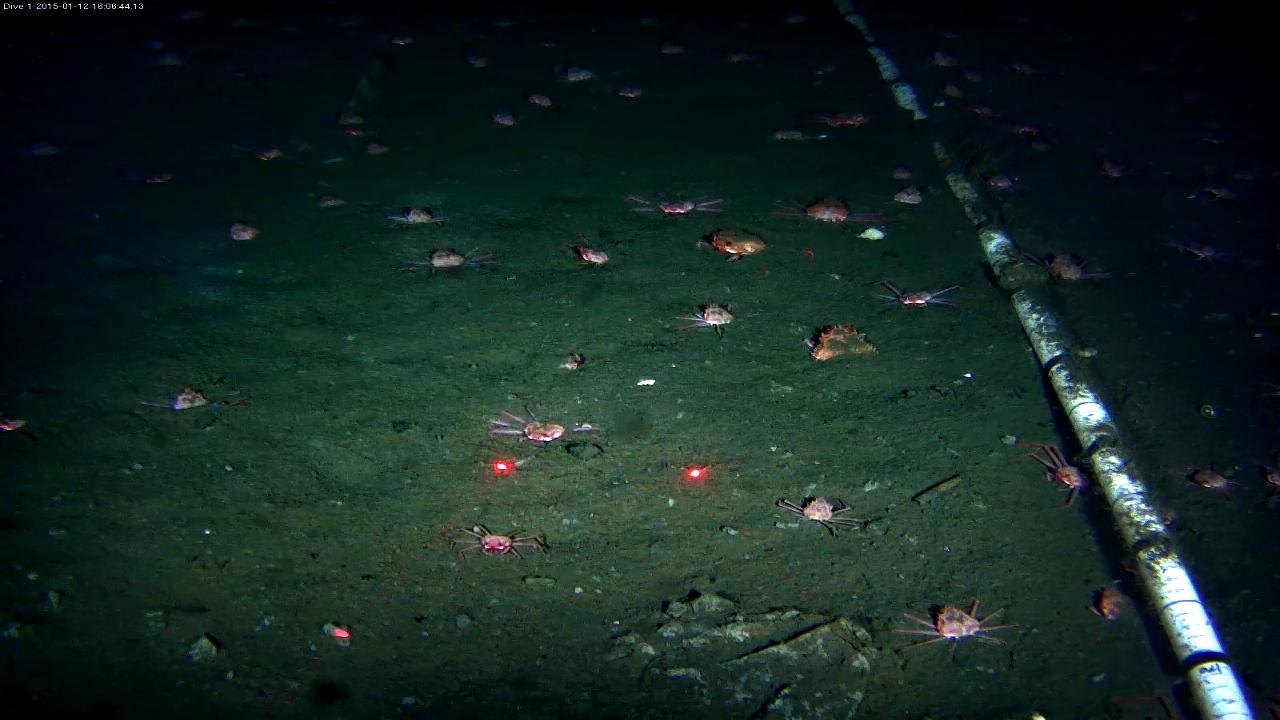}
\includegraphics[width=0.195\textwidth]{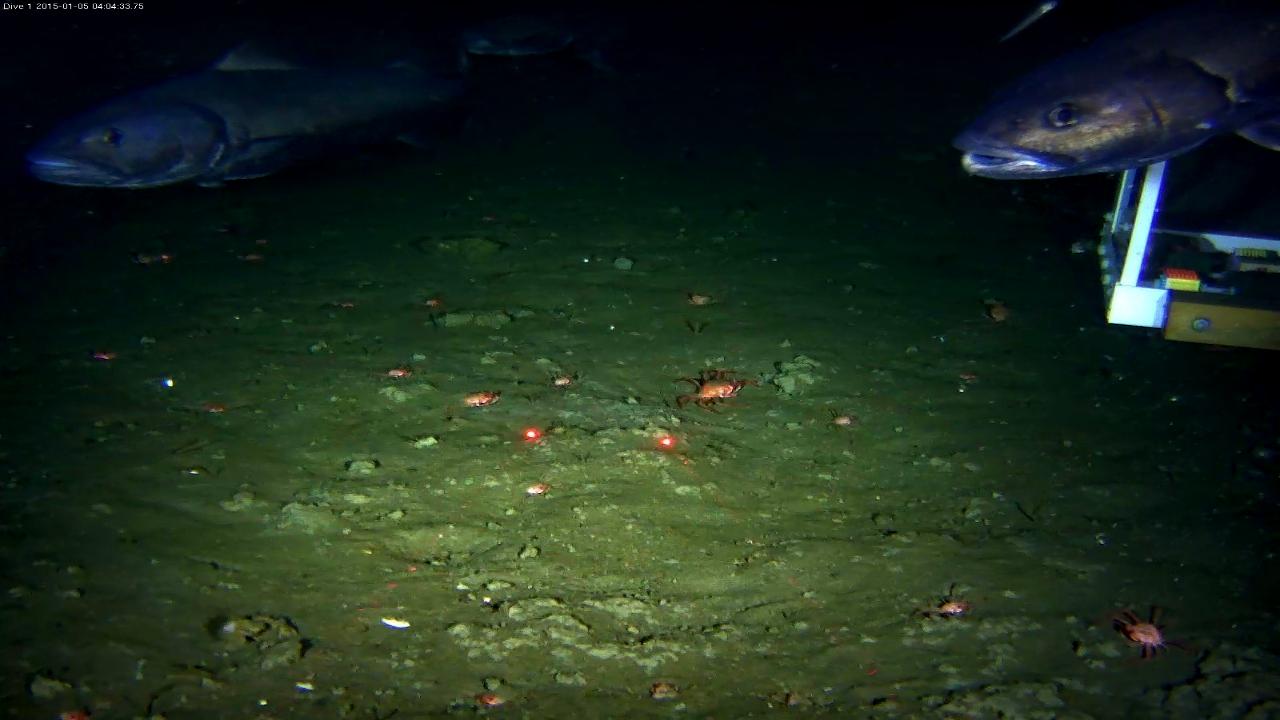}
\includegraphics[width=0.195\textwidth]{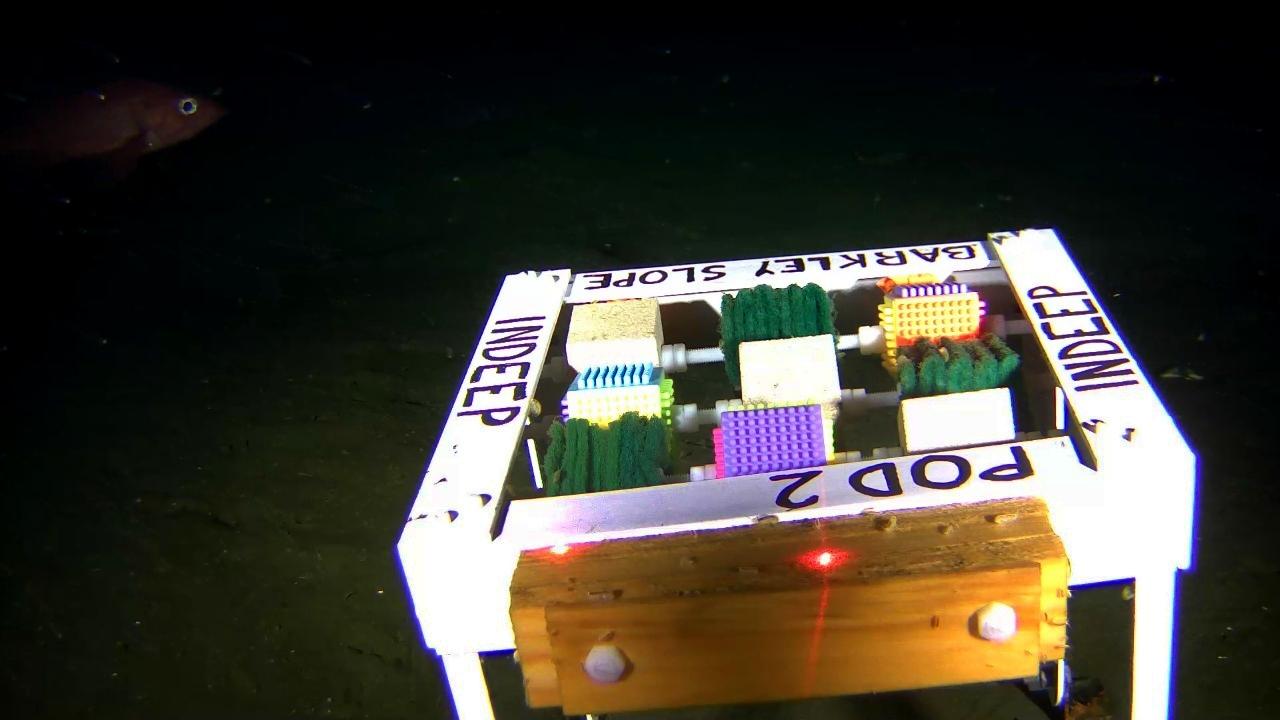}
\includegraphics[width=0.195\textwidth]{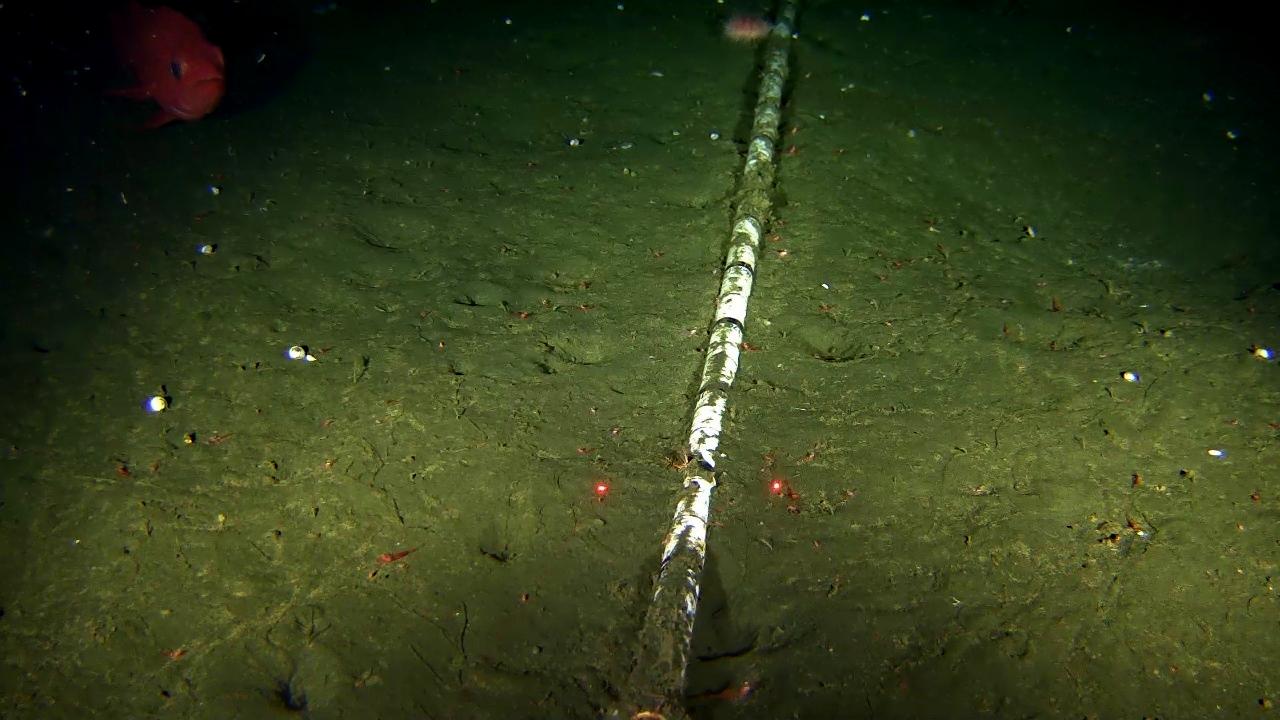}
\end{minipage}\hfill

\begin{minipage}[c]{0.095\textwidth}
\textbf{Zhang}~\cite{zhang2017fast} \label{fig:experiment_zhang}
\end{minipage}%
\begin{minipage}[c]{0.9\linewidth}
\includegraphics[width=0.195\textwidth]{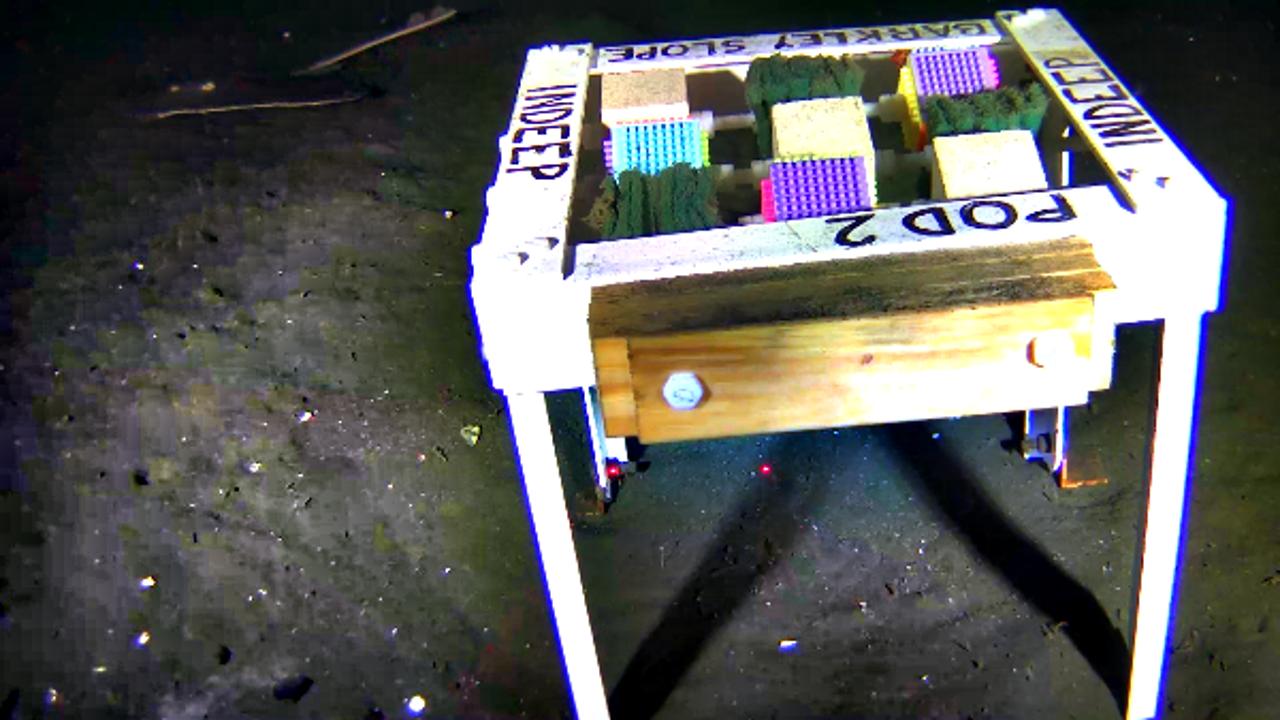}
\includegraphics[width=0.195\textwidth]{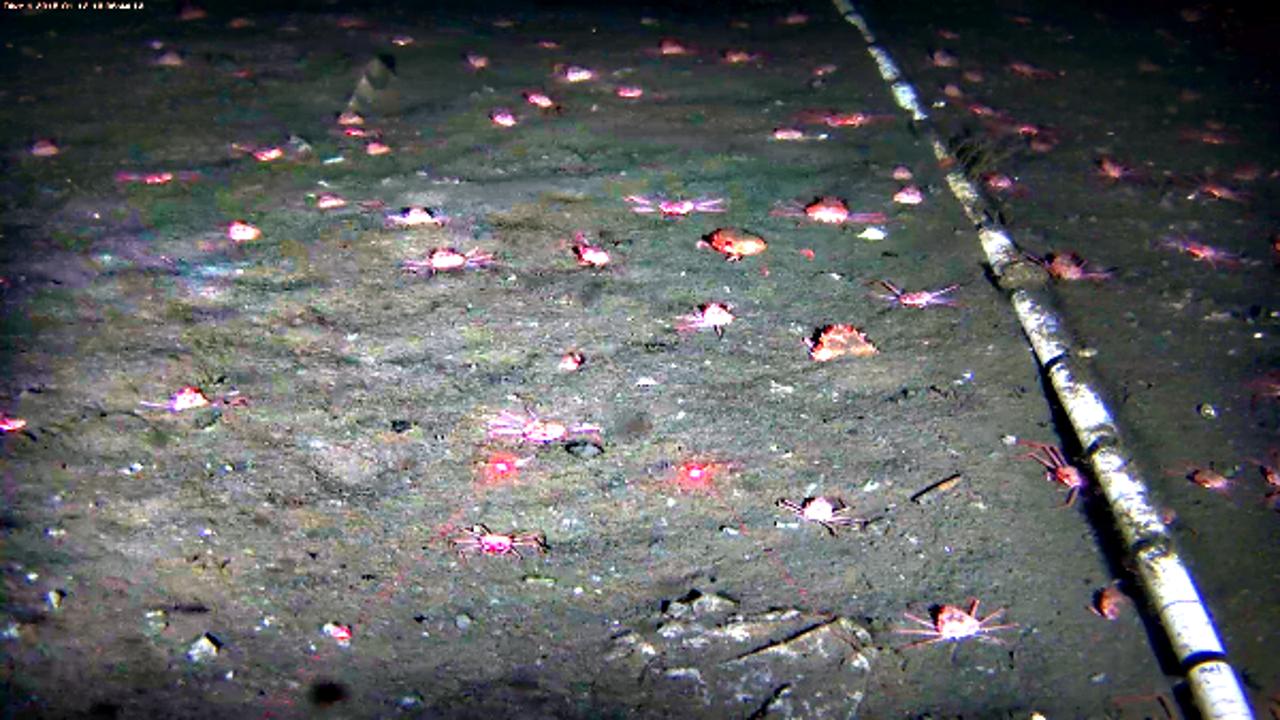}
\includegraphics[width=0.195\textwidth]{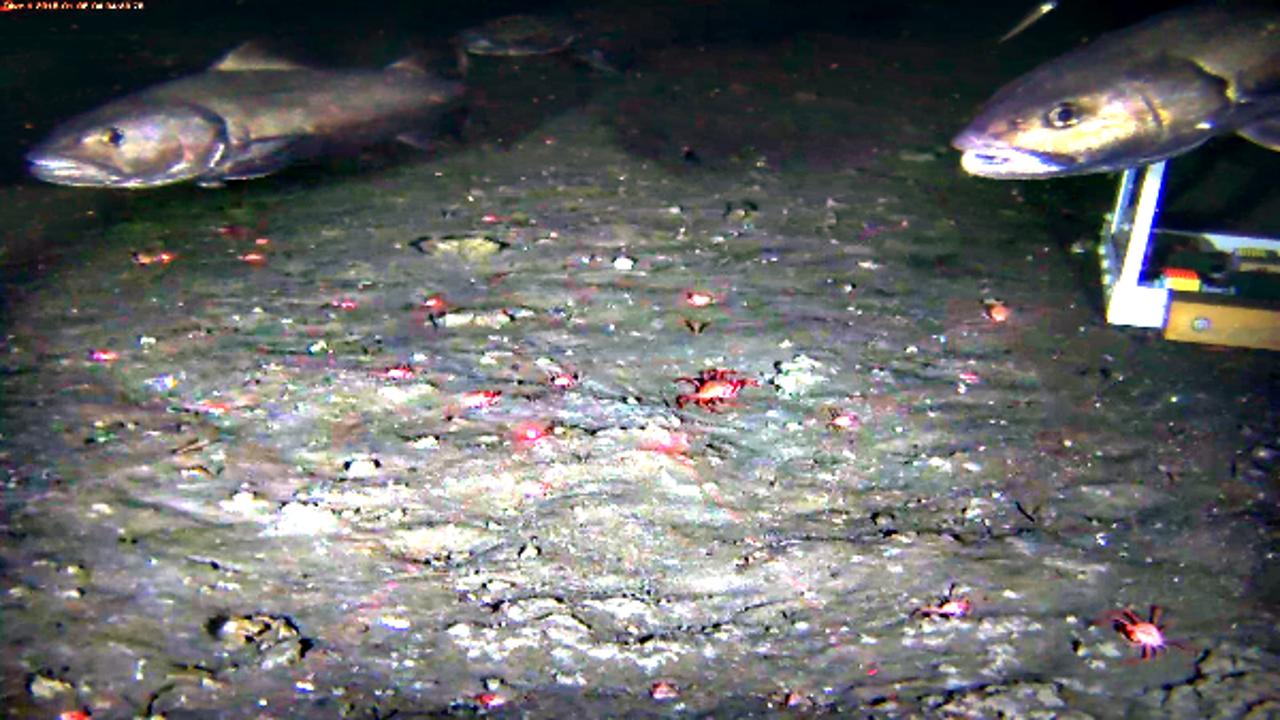}
\includegraphics[width=0.195\textwidth]{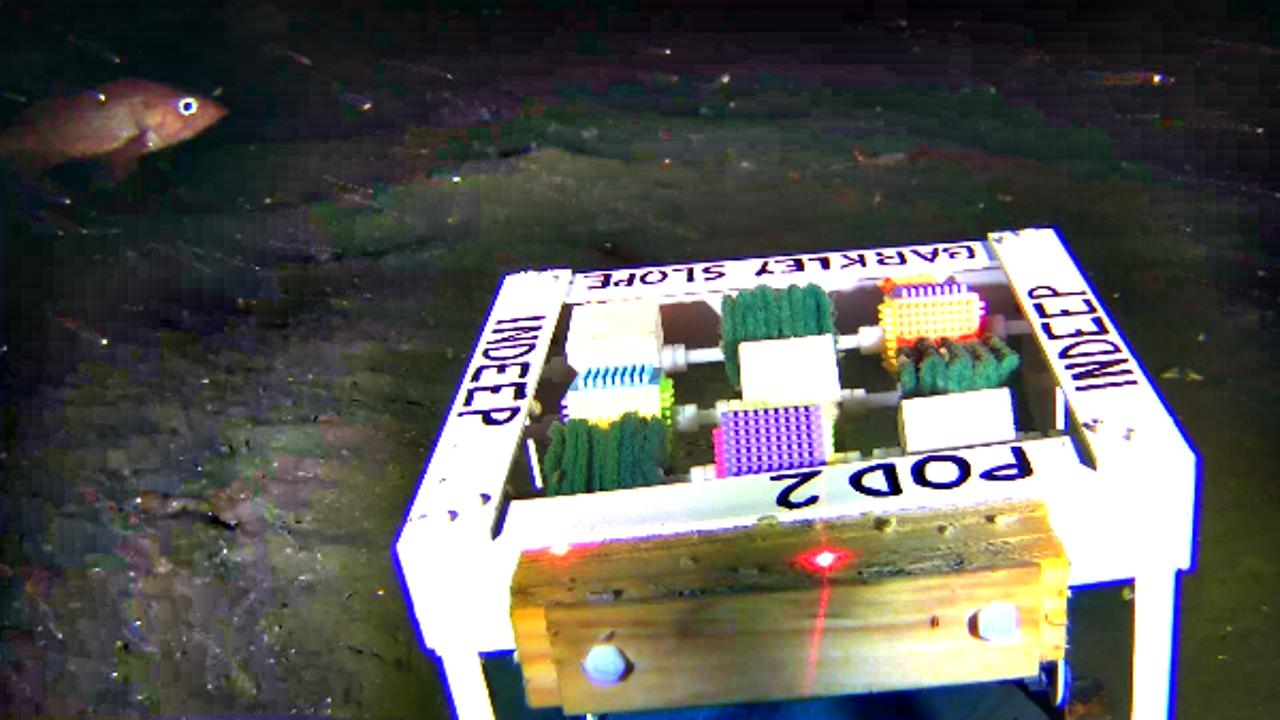}
\includegraphics[width=0.195\textwidth]{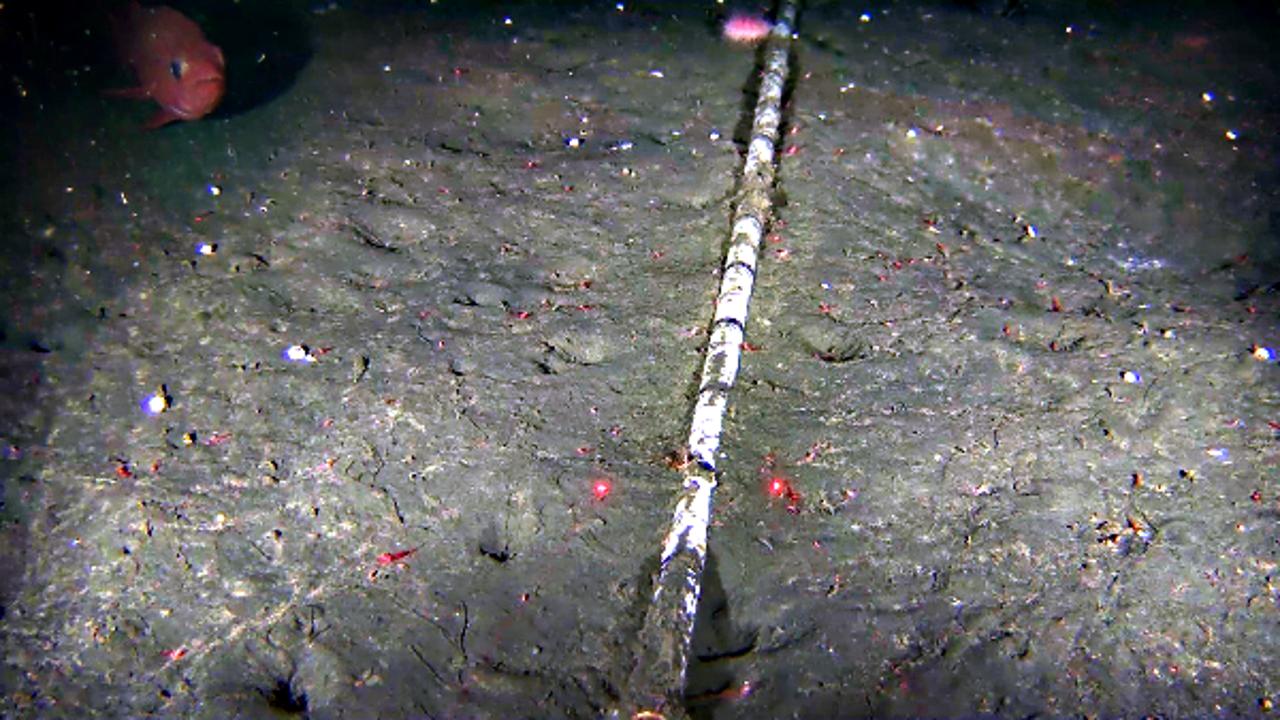}
\end{minipage}\hfill

\begin{minipage}[c]{0.095\textwidth}
\textbf{Guo}~\cite{guo2016lime} \label{fig:experiment_guo}
\end{minipage}%
\begin{minipage}[c]{0.9\linewidth}
\includegraphics[width=0.195\textwidth]{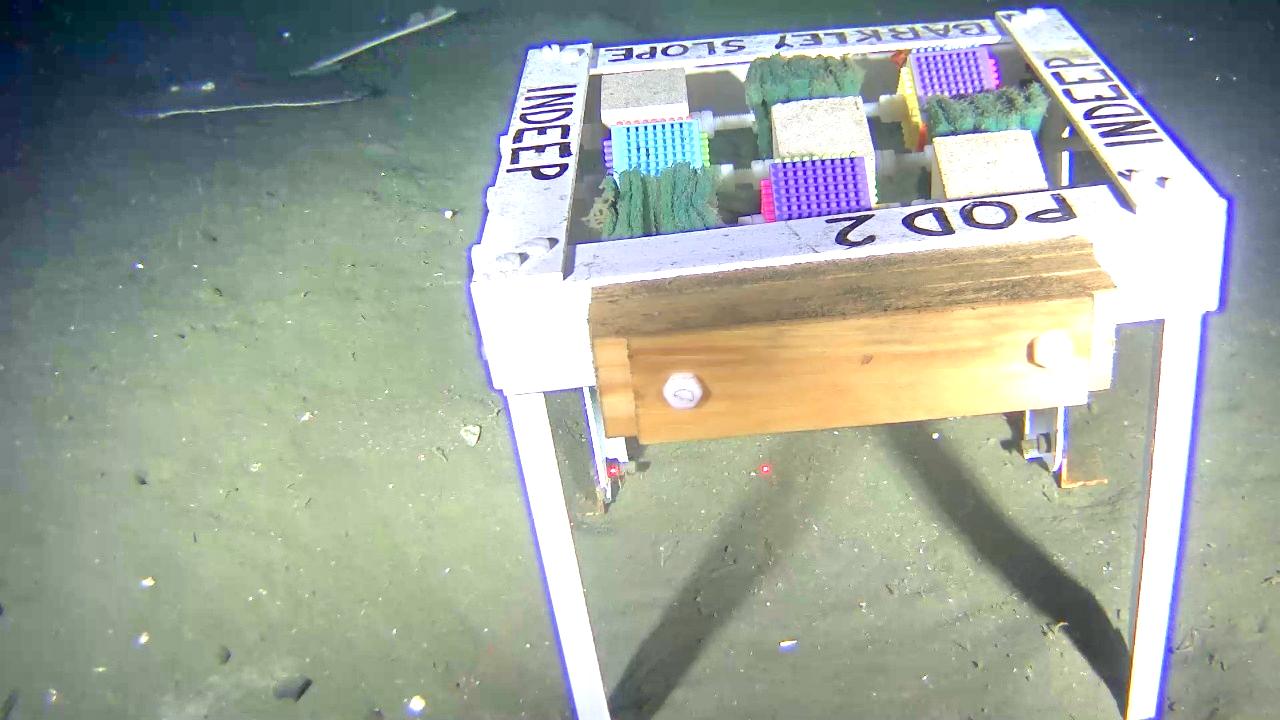}
\includegraphics[width=0.195\textwidth]{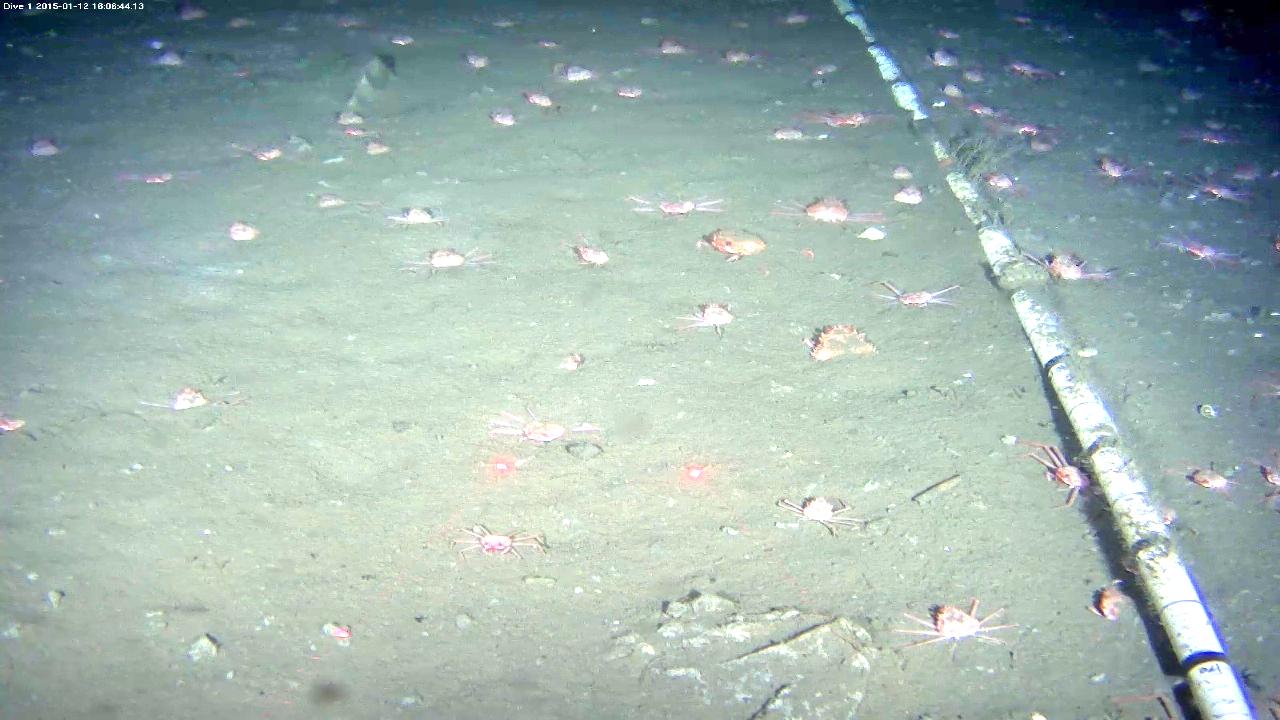}
\includegraphics[width=0.195\textwidth]{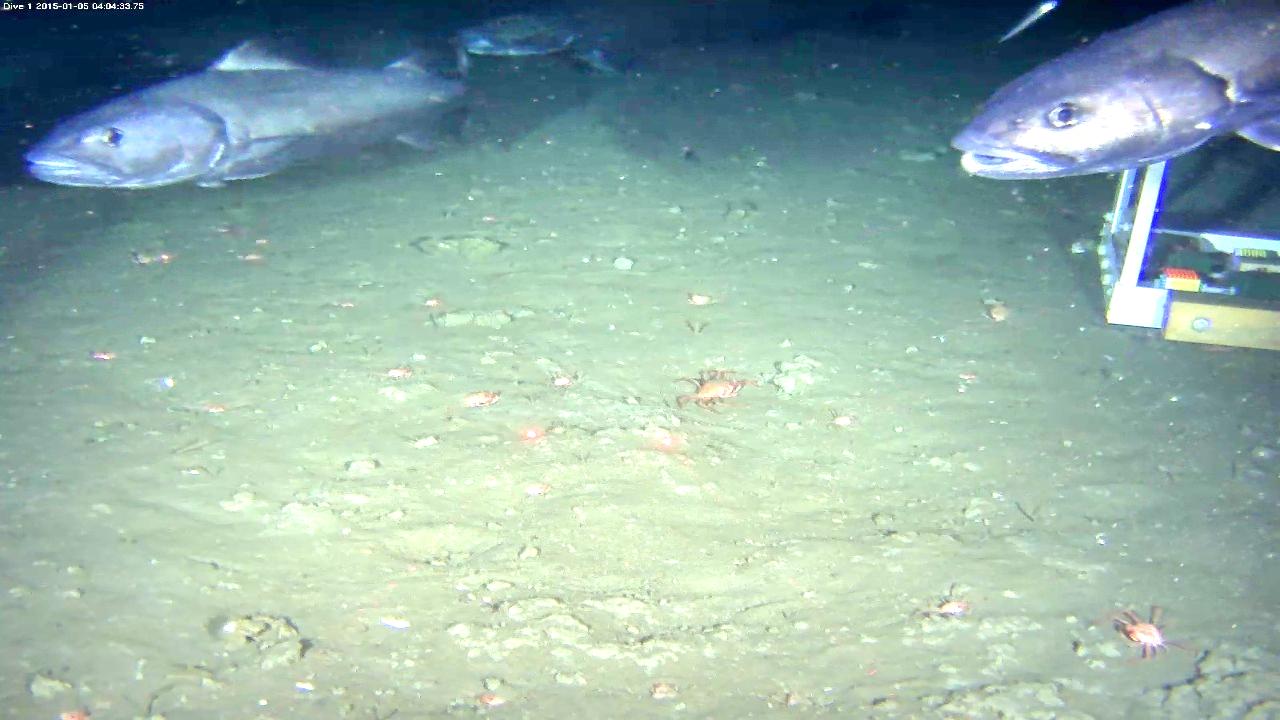}
\includegraphics[width=0.195\textwidth]{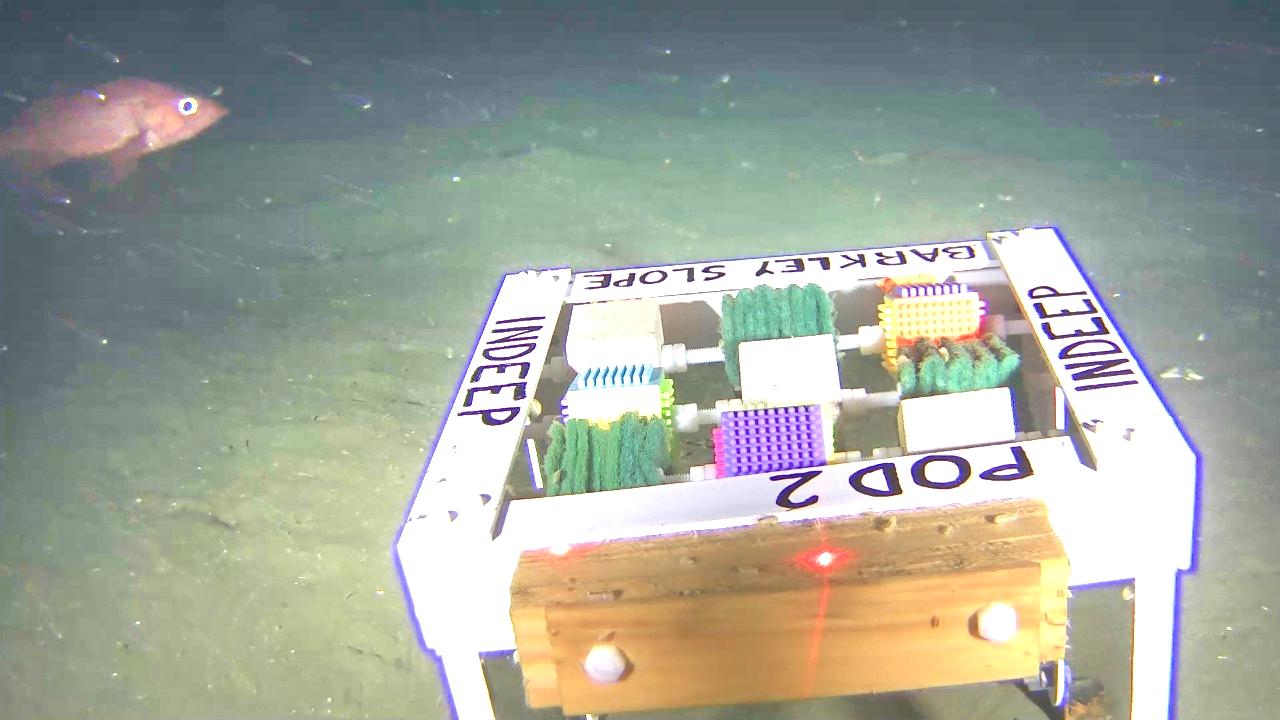}
\includegraphics[width=0.195\textwidth]{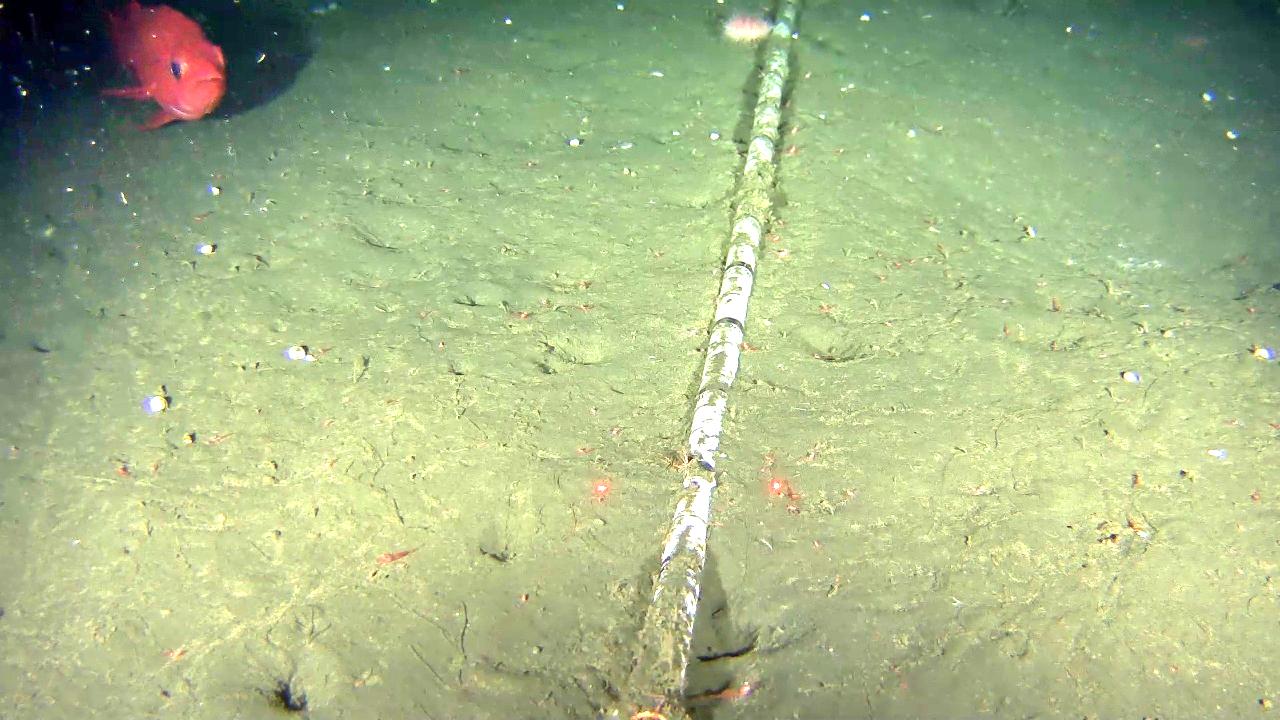}
\vspace{-2em}
\end{minipage}\hfill
\end{center}
\caption{Output of diverse image enhancement methods on samples from the OceanDark dataset~\cite{mdpipaper}.}
\label{fig:experiment}
\end{figure*}

Experimental results show the capacity of the proposed method of drastically reducing low-light regions of inputs without creating washed-out outputs (see Figure \ref{fig:experiment}). Although other methods can be used for the enhancement of similar scenarios with remarkable results, our experiments show that \luwe~outperforms the current state-of-the-art approaches in the task of enhancing low-light underwater images. Our proposed contrast-guided computational pipeline is also expected to work well in other mediums, such as night-time and low-light aerial scenes.\par
Future developments will focus on the adaptation of \luwe~for aerial low-light images, its evaluation in additional datasets, and the use of data-driven systems in the framework's computational pipeline (e.g., a CNN-based network for the estimation of atmospheric lighting models).

\begin{table*}[ht]
\begin{center}
\begin{tabular}{p{0.11\linewidth}p{0.095\linewidth}p{0.1\linewidth}p{0.1\linewidth}p{0.1\linewidth}p{0.1\linewidth}p{0.1\linewidth}p{0.1\linewidth}}
\hline
Method & UIQM~\cite{panetta2015human} & PCQI~\cite{wang2015patch} & GCF~\cite{matkovic2005global} & $e$-score~\cite{hautiere2011blind} & $r$-score~\cite{hautiere2011blind}
 & FADE~\cite{choi2015referenceless} & SURF~\cite{bay2008speeded}  \\
\hline\hline

Original & $0.88 \pm 0.13$ & \centering $1$ & $3.28 \pm 0.62$ & \centering N/A & \centering $1$ & $1.91 \pm 0.79$ & $340 \pm 293$ \\
Marques~\cite{mdpipaper} & $0.99 \pm 0.12$ & $0.85 \pm 0.03$ & $3.41 \pm 0.71$ & $0.28 \pm 0.32$ & $2.75 \pm 0.76$ & $0.46 \pm 0.18$ & $705 \pm 470$ \\
Berman~\cite{berman2017diving} & $1.00 \pm 0.18$ & $0.78 \pm 0.15$ & $3.84 \pm 1.07$ & $0.25 \pm 0.50$ & $2.91 \pm 1.96$ & $1.15 \pm 0.40$ & $425 \pm 317$ \\
Fu~\cite{fu2014retinex} & $0.93 \pm 0.15$ & $0.93 \pm 0.09$ & $3.28 \pm 0.57$ & $0.09 \pm 0.39$ & $1.72 \pm 0.35$ & $1.75 \pm 0.25$ & $865 \pm 478$ \\
Cho~\cite{cho2017visibility} & $1.24 \pm 0.15$ & $0.87 \pm 0.04$ & $4.11 \pm 0.84$ & $0.89 \pm 0.54$ & $1.72 \pm 0.09$ & $1.81 \pm 0.52$ & $751 \pm 428$ \\
Drews~\cite{drews2013transmission} & $1.38 \pm 0.14$ & $0.85 \pm 0.05$ & $4.70 \pm 0.89$ & $1.06 \pm 0.83$ & $1.29 \pm 0.31$ & $2.08 \pm 0.90$ & $589 \pm 324$ \\
Zhang~\cite{zhang2017fast} & $1.28 \pm 0.08$ & $1.03 \pm 0.07$ & $\textbf{6.34} \pm \textbf{0.74}$ & $1.48 \pm 0.88$ & $3.70 \pm 1.00$ & $0.72 \pm 0.39$ & $1719 \pm 620$ \\
Guo~\cite{guo2016lime} & $0.93 \pm 0.13$ & $0.86 \pm 0.03$ & $3.50 \pm 0.73$ & $0.16 \pm 0.14$ & $2.21 \pm 0.64$ & $0.60 \pm 0.24$ & $607 \pm 452$ \\
\luwe~ & $\textbf{1.38} \pm \textbf{0.11}$ & $\textbf{1.17} \pm \textbf{0.07}$ & $4.89 \pm 0.66$ & $\textbf{1.82} \pm \textbf{0.83}$ & $\textbf{4.61} \pm \textbf{0.58}$ & $\textbf{0.42} \pm \textbf{0.20}$ & $\textbf{1856} \pm \textbf{655}$ \\
\hline
\end{tabular}
\vspace{-1.5em}
\end{center}
\caption{Mean and standard deviation of seven metrics computed on all samples of the OceanDark dataset~\cite{mdpipaper}. The results compare the output of diverse image enhancing methods. Best results (i.e., higher, with the exception of darkness-indicating FADE~\cite{choi2015referenceless}) are bolded.}
\label{tab:results}
\end{table*}

{\small
\bibliographystyle{ieee_fullname}
\bibliography{final}
}

\end{document}